\pdfoutput=1

\documentclass[11pt]{article}

\usepackage{acl}

\usepackage{times}
\usepackage{latexsym}

\usepackage[T1]{fontenc}

\usepackage[utf8]{inputenc}

\usepackage{microtype}

\usepackage{inconsolata}

\usepackage{amsmath,amsfonts,bm}
\usepackage{amsthm}

\usepackage{amssymb,mathtools}
\usepackage{bbm}
\usepackage{nccmath}
\usepackage{comment}

\DeclareMathOperator{\btheta}{\bm{\theta}}

\DeclareMathOperator{\bx}{\mathbf{x}}

\DeclareMathOperator{\bz}{\mathbf{z}}

\usepackage{subcaption}
\usepackage{booktabs}
\usepackage{multirow}
\usepackage{arydshln}
\usepackage{algorithm}
\usepackage{algpseudocode}
\algnewcommand{\LineComment}[1]{\State \(\triangleright\) #1}

\usepackage{fontawesome5}

\usepackage{tikz}
\usepackage{tikz-dependency}
\newcommand*\circled[1]{\tikz[baseline=(char.base)]{
            \node[shape=circle,draw,inner sep=.6pt] (char) {#1};}}

\usepackage[noabbrev,capitalize,nameinlink]{cleveref}

\usepackage{adjustbox}

\title{Non-Exchangeable Conformal Language Generation with\\Nearest Neighbors}

 \author{Dennis Ulmer\textsuperscript{ \faCompass, \faFlag} \hspace{.5em}  Chrysoula Zerva\textsuperscript{ \faPhone, \faFortAwesome} \hspace{.5em} André F.T. Martins\textsuperscript{ \faPhone, \faFortAwesome, \faComments}\\
       \textsuperscript{\faCompass} IT University of Copenhagen,  
       \textsuperscript{\faFlag} Pioneer Centre for Artificial Intelligence,\\
       \textsuperscript{\faPhone} Instituto de Telecomunicações, \textsuperscript{\faComments} Unbabel,\\
       \textsuperscript{\faFortAwesome} Instituto Superior Técnico, Universidade de Lisboa (Lisbon ELLIS Unit)\\
        \texttt{dennis.ulmer@mailbox.org}}

\begin{document}
\maketitle
\begin{abstract}
    Quantifying uncertainty in automatically generated text is important for letting humans check potential hallucinations and making systems more reliable. 
    Conformal prediction is an attractive framework to provide predictions imbued with statistical guarantees,
    however, its application to text generation is challenging since any i.i.d.\@ assumptions are not realistic.
    In this paper, we bridge this gap by leveraging recent results on \textit{non-exchangeable} conformal prediction, which still ensures bounds on coverage. The result, \emph{non-exchangeable conformal nucleus sampling}, is a novel extension of the conformal prediction framework to generation based on nearest neighbors. Our method can be used post-hoc for an arbitrary model without extra training and supplies token-level, calibrated prediction sets equipped with statistical guarantees.
    Experiments in machine translation and language modeling show encouraging results in generation quality.
    By also producing tighter prediction sets with good coverage, we thus give a more theoretically principled way to perform sampling with conformal guarantees.
\end{abstract}

\section{Introduction}

Natural language generation (NLG) is a multi-faceted field spanning applications such as machine translation (MT), language modeling (LM), summarization, question answering and dialogue generation. Owing to the recent success of large language models (LLMs) such as GPT-4 \citep{openai2023gpt4}, BLOOM \citep{scao2022bloom} or LLaMA \citep{touvron2023llama}, natural language modeling with stochastic decoding (sampling) is increasingly used as an interface with end users. 
While sampling allows for more fluent and varied text, few methods exist to evaluate the reliability of generated text and adequacy of the underlying sampling method. 
This is particularly relevant for generation scenarios where pre-trained models are applied to new data with potentially different distribution to the training data, increasing the risk of generating erroneous, misleading, and potentially harmful text \cite{ji2023survey, guerreiro2023looking, pan2023risk, alkaissi2023artificial, azamfirei2023large}.\\

\begin{figure}[tb!]
    \centering 
    \includegraphics[width=1.05\columnwidth]{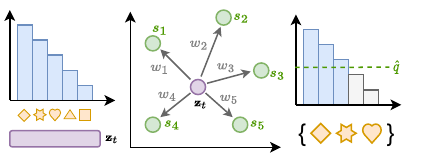}
    \caption{Schematic representation of our approach. A decoder hidden representation $\bz_t$ is used during inference to retrieve the nearest neighbors and their non-conformity scores $s_k$. Their relevance is determined by using their distance to compute weights $w_k$, resulting in the quantile $\hat{q}$ that forms conformal prediction sets.}\label{fig:schematic}
\end{figure}

Conformal prediction \citep{vovk2005algorithmic, papadopoulos2002inductive, angelopoulos2021gentle} has recently gained popularity by providing calibrated prediction sets that are imbued with statistical guarantees about containing the correct solution. Nevertheless, applying conformal prediction to NLG is not trivial and comes with a major obstacle: The conditional generation process breaks the independence and identical distribution (i.i.d.)\@ assumption underlying conformal prediction techniques. We tackle this problem by drawing inspiration from recent advances in nearest neighbor language modeling \citep{khandelwal2020generalization, he2023efficient, xu2023nearest} and machine translation \citep{khandelwal2020nearest, zheng2021adaptive, meng2022fast, henrique2022chunk}. This way, we are able to dynamically generate calibration sets during inference that are able to maintain statistical guarantees. We schematically illustrate non-exchangeable conformal nucleus sampling in \cref{fig:schematic}:
In the first step, we obtain a (sorted) probability distribution over tokens and a latent representation $\bz_t$ for the current generation step from the model.
In a second step, we use the latent representation to query a datastore for similar, previously stored representations and their corresponding non-conformity scores, $s_k$.
These scores are then used to compute a threshold $\hat{q}$ based on the theory of non-exchangeable conformal prediction \citep{barber2022conformal}, which defines a smaller set of tokens that is sampled from.\footnote{For simplicity, the figure depicts the simplest form of prediction sets used in conformal prediction. In practice, we use the adaptive prediction sets explained in \cref{sec:method}.}


\paragraph{Contributions.} We present a general-purpose extension of the conformal framework to NLG by tackling the problems above. Our contributions are as follows: \circled{1} To the best of our knowledge, we are the first to present a novel technique based on \textit{non-exchangeable} conformal prediction and to apply it to language generation to produce calibrated prediction sets. \circled{2} We validate the effectiveness of the method in a Language Modeling and Machine Translation context, evaluating the coverage of the calibrated prediction sets and showing that our method is on par with or even outperforms other sampling-based techniques in terms of generation quality, all while maintaining tighter prediction sets and better coverage. \circled{3} We finally demonstrate that these properties are also maintained under distributional shift induced by corrupting the model's latent representations. \circled{4} We publish all the code for this project in an open-source repository.\footnote{\url{https://github.com/Kaleidophon/non-exchangeable-conformal-language-generation}.}

\section{Related Work}

\paragraph{Conformal Prediction.} Conformal prediction is a line of work that has recently regained interest in machine learning by producing prediction sets with certain statistical guarantees about containing the correct prediction \citep{vovk2005algorithmic, papadopoulos2002inductive, angelopoulos2021gentle}.  
As the size of prediction sets is calibrated to fulfill these guarantees, one can also see the size of the prediction set itself as a proxy of the uncertainty of a model---the larger the set, the more possible predictions have to be included in order to maintain the coverage guarantee.
Conformal prediction has already found diverse applications in NLP for classification  \citep{maltoudoglou2020bert, fisch2021few, schuster2021consistent, fisch2022conformal, choubey-etal-2022-conformal, kumar2023conformal} and sequence labeling problems \citep{dey2021conformal}, as well as quality estimation \citep{giovannotti2023evaluating, zerva2023conformalizing}.
Unfortunately, generation problems are challenging due to their sequential nature and constant breaking of the i.i.d.\@ assumption, so existing works operate on the sequence-level instead \citep{quach2023conformal, ren2023robots, deutschmann2023conformal}.
Conformal procedures for time-series \citep{xu2021conformal, lin2022conformal, oliveira2022split, zaffran2022adaptive} and general non-i.i.d.\@ data \citep{tibshirani2019conformal, barber2022conformal, guan2023localized, farinhas2024nonexchangeable} have been proposed in the literature.
The most related work to ours is given by \citet{ravfogel2023conformal}, who apply the standard conformal prediction setup to NLG, arguing that Markov chains are a type of $\beta$-mixing processes, for which \citet{oliveira2022split} showed coverage to degrade by an only negligible amount. However, \citeauthor{ravfogel2023conformal} do not investigate this claim empirically, and furthermore do not find any benefits when generating sequences.
In another related work, \citet{quach2023conformal} propose an approach that is specifically tailored toward language modeling. 
However, their prediction sets contain entire sequences instead of single tokens. In contrast, our token-level prediction sets are useful for constraining the options during generation and their widths can represent model uncertainty.

\paragraph{Uncertainty in NLP.} Modeling uncertainty in NLP has already been studied in classification \citep{van2022benchmarking, ulmer-etal-2022-exploring, holm2022revisiting} and regression settings \citep{beck-etal-2016-exploring, glushkova-etal-2021-uncertainty-aware, zerva2022dis}.
However, NLG proves more challenging due to it non-i.i.d.\@ and combinatorial nature.
Some works have proposed Bayesian Deep Learning methods for NLG: \citet{xiao2020wat} use Monte Carlo Dropout \citep{gal2016dropout} to produce multiple generations for the same input and measure their pair-wise BLEU scores. \citet{malinin2020uncertainty} define extensions of mutual information for structured prediction.
Other existing approaches try to account for the paraphrastic nature of language by modeling the entropy over meaning classes \citep{kuhn2023semantic}, investigate the use of linguistic markers to indicate uncertainty \citep{zhou2023navigating} or ask the model directly for its confidence \citep{lin2022teaching, kadavath2022language}.
\citet{baan2023uncertainty} provide an extensive overview of the theory and current state of the field.

\section{Background}\label{sec:background}

\paragraph{Conformal Prediction.} Conformal prediction is an attractive method for uncertainty quantification due to its statistical coverage guarantees \citep{vovk2005algorithmic, papadopoulos2002inductive, angelopoulos2021gentle}. Given some predictor, a held-out calibration set $\{(\bx_i, y_i)\}_{i=1}^N$, and a pre-defined miscoverage level $\alpha$ (e.g., 0.1), the calibration set is used to obtain \textit{prediction sets} $\mathcal{C}(\bx^*)$ for a new test point $\bx^*$ satisfying 
\begin{equation}\label{eq:conformal-prediction-guarantees}
    p\Big(y^* \in \mathcal{C}(\bx^*)\Big) \ge 1 - \alpha,
\end{equation}
that is, the probability of the prediction set $\mathcal{C}(\bx^*)$  containing the correct label $y^*$ is \textit{at least} $1-\alpha$. 
This is achieved by the following recipe: 
Firstly, one has to define a \emph{non-conformity score}, that provides an estimate of the distance of the test point to the rest of the data, i.e., a proxy for the uncertainty over the test point predictions. 
In this context, the score can be as simple as $s_i = 1 - p_{\btheta}(y|\bx)$, i.e.\@ one minus the softmax probability of the true class, which will be higher when the model is wrong or less confident.
Next, we define $\hat{q}$ as the $\big\lceil(N+1)(1 - \alpha)/N\big\rceil$-th quantile of the non-conformity scores. 
Then, when we make a new prediction for a test point $\bx^*$, we can create prediction sets defined as

\begin{equation}\label{eq:conformal-prediction-sets}
    \mathcal{C}(\bx^*) = \Big\{y\ \Big|\ p_{\btheta}(y|\bx^*) \ge 1 - \hat{q}\Big\},
\end{equation}

\noindent which is guaranteed to fulfil the coverage requirement in \cref{eq:conformal-prediction-guarantees} for i.i.d.\@ data \citep{vovk2005algorithmic, angelopoulos2021gentle}.

\paragraph{Non-exchangeable Conformal Prediction.} \citet{barber2022conformal} address a major shortcoming in the method above: When a test point and the calibration data are not i.i.d.,%
\footnote{In fact, the coverage guarantee applies to the case where the data is \textit{exchangeable}, a weaker requirement than i.i.d. Specifically, a series of random variables is exchangeable if their joint distribution is unaffected by a change of their order.} %
the distributional drift causes any previously found $\hat{q}$ to be miscalibrated, and thus the intended coverage can no longer be guaranteed. However, we can still perform conformal prediction by assigning a weight $w_i \in [0, 1]$ to every calibration data point, reflecting its relevance---i.e.\@ assigning lower weights to points far away from the test distribution. Then, by normalizing the weights with $\tilde{w}_i = w_i / (1 + \sum_{i=1}^N w_i)$, we define the quantile as

\begin{equation}\label{eq:non-exchangeable-quantile}
    \hat{q} = \inf \Big\{q\ \Big| \sum_{i=1}^N \tilde{w}_i \mathbf{1}\big\{s_i\le q \big\} \ge 1 - \alpha \Big\},
\end{equation}

\noindent with $\mathbf{1}\{\cdot\}$ denoting the indicator function. 
The construction of the prediction sets then follows the same steps as before.
Most notably, the coverage guarantee in \cref{eq:conformal-prediction-guarantees} now changes to 

\begin{equation}\label{eq:non-exchangeable-conformal-prediction-guarantees}
    p\Big(y^* \in \mathcal{C}(\bx^*)\Big) \ge 1 - \alpha - \sum_{i=1}^N \tilde{w}_i\varepsilon_i,
\end{equation}

\noindent with an extra term including the \textit{total variation distance} between the distribution of a calibration and a test point, $\varepsilon_i = d_\text{TV}\big((\bx_i, y_i), (\bx^*, y^*)\big)$.%
\footnote{In this expression, $(\bx_i, y_i)$ and $(\bx^*, y^*)$ denote random variables and the total variation distance is between the two underlying distributions. See \citet{barber2022conformal} for details.} %
Unfortunately, this term is hard to estimate or bound, nevertheless, the selection of appropriate weights that can capture the relevance of calibration points to the test set should moderate both the impact of the distant data points on the estimation of the prediction set and the impact of $d_{\mathrm{TV}}$ on the coverage bound. 
In other words, for large $d_{\mathrm{TV}}$ values we expect to have smaller weights, that allow us to achieve coverage close to the desired values.
We show in our experiments that the loss of coverage when using nearest neighbor weights is limited and revisit the practical implications in \cref{sec:discussion}.

\subsection{Method: Non-exchangeable Conformal Language Generation through Nearest Neighbors}\label{sec:method}

We now present a novel method to apply conformal prediction in NLG by synthesizing the non-exchangeable approach of \citet{barber2022conformal} with $k$-NN search-augmented neural models \citep{khandelwal2020nearest, khandelwal2020generalization}.
The related approach by \citet{ravfogel2023conformal} calibrates prediction sets within bins of similar entropies using the non-exchangeable procedure described in \cref{sec:background}. 
However, this implies that we would use semantically unrelated (sub-)sequences to calibrate the model---in fact, we show experimentally that this approach obtains generally trivial coverage by producing extremely wide prediction sets. Instead, we propose to perform a \emph{dynamic} calibration step during model inference, only considering the most relevant data points from the calibration set. We do this in the following way: Given a dataset $\{(\bx^{(i)}, y^{(i)})\}$ of sequences $\bx^{(i)} = (\bx_1^{(i)}, \ldots, \bx_S^{(i)})$ and corresponding references consisting of gold tokens $y^{(i)} = (y_1^{(i)}, \ldots, y_T^{(i)})$, we extract the model's decoder activations $\bz_t^{(i)} \in \mathbb{R}^d$ and conformity scores $s_t^{(i)}$.\footnote{In this phase, we do not let the model generate freely, but feed it the gold prefix during the decoding process to make sure that conformity scores can be computed correctly.} We save those in a datastore allowing for fast and efficient nearest neighbor search using FAISS \citep{johnson2019billion}. 
In the inference phase, during every decoding step, we then use the decoder hidden state $\bz_t^*$ to query the data store for the $K$ nearest neighbors and their conformity scores and record their distances. We use the squared $l_2$ distance to compute the weight $w_k$ as 

\begin{equation}\label{eq:weight-equation}
    w_k = \exp\Big(-\big|\big|\bz_t^* - {\bz_k}\big|\big|^2_2\ /\ \tau\Big),
\end{equation}

\noindent where $\tau$ corresponds to a temperature hyperparameter.\footnote{Using this formulation of the weights $w_k$ that depends on the data deviates from the assumptions of original proof, as discussed in \citet{barber2022conformal}, §4.5.
Nevertheless, our results in \cref{sec:experiments} and those by  \citet{farinhas2024nonexchangeable} show that the obtained bound in \cref{eq:non-exchangeable-conformal-prediction-guarantees} still remains useful.}
This formulation is equivalent to a RBF kernel with scale parameter $\tau$.
Finally, we use the weights to compute the quantile $\hat{q}$ as in \cref{eq:non-exchangeable-quantile}.
The entire algorithm is given in \cref{app:algorithm}.

\paragraph{Adaptive Prediction Sets.} The efficacy of conformal prediction hinges on the choice of non-conformity score, with the simple non-conformity score $s_i = 1 - p_{\btheta}(y_t|\bx, y_{<t})$ known to undercover hard and overcover easy subpopulations of the data.
Due to the diverse nature of language, we therefore opt for \emph{adaptive prediction sets} \citep{angelopoulos2020uncertainty, romano2020classification}. 
Adaptive prediction sets redefine the non-conformity score as the cumulative probability over classes (after sorting descendingly) necessary to reach the correct class.
Intuitively, this means that we included all classes whose cumulative probability does not surpass $\hat{q}$.
Compared to the simple conformity score, this produces wider predictions sets for hard inputs, encompassing more potentially plausible continuations in a language context.
A more formal definition  is given in \cref{app:adaptive-prediction-sets}.

\section{Experiments}\label{sec:experiments}

In the following sections, we conduct experiments in both language modeling and machine translation.
For machine translation we opt for the 400 million and 1.2 billion parameter versions of the M2M100 model \citep{fan2021beyond} on the WMT-2022 shared task datasets for German to English and Japanese to English \citep{kocmi2022findings}.
For Language Modelling, we use the 350 million and 1.3 billion parameter versions of the OPT model \citep{zhang2022opt} and replicate the setup by \citet{ravfogel2023conformal}:
We calibrate our model on $10000$ sentences from a 2022 English Wikipedia dump \citep{wikidump} and test coverage and generation on $1000$ sentences from OpenWebText \citep{Gokaslan2019OpenWeb}.\footnote{Data obtained through the Hugging Face \texttt{datasets} package \citep{lhoest2022datasets}: \url{https://huggingface.co/datasets/wikipedia} and \url{https://huggingface.co/datasets/stas/openwebtext-10k}.}
All models are used in a zero-shot setup \emph{without extra training or finetuning}.
For the datastore, we use the implementation by FAISS library \citep{johnson2019billion}, computing $2048$ clusters in total and probing $32$ clusters per query. 
We also summarize the environmental impact of our experiments in \cref{app:environmental-impact}.

\subsection{Evaluating Coverage}\label{sec:retrieval-quality}

\begin{table*}[htb!]
    \centering
    \resizebox{0.978\textwidth}{!}{
    \renewcommand{\arraystretch}{1.4}
    \begin{tabular}{@{}lllrccccrcccc@{}}
        \toprule[0.05cm]
         & & & \multicolumn{5}{c}{de $ \rightarrow $ en} & \multicolumn{5}{c}{ja $ \rightarrow $ en} \\
         \cmidrule(lr){4-8}\cmidrule(lr){9-13} 
         & Method & Dist. & $\tau$ & \% \textsc{Coverage} & $\varnothing$ \textsc{Width} $\downarrow$ & \textsc{Scc} $\uparrow$ & \textsc{Ecg} $\downarrow$ & $\tau$ & \% \textsc{Coverage} & $\varnothing$ \textsc{Width} $\downarrow$ & \textsc{Scc} $\uparrow$ & \textsc{Ecg} $\downarrow$ \\
        \midrule 
        \multirow{5}{*}{\rotatebox[origin=c]{90}{M2M100$_\text{(400M)}$}} & Nucleus Sampling & - & - & $0.9207$ & $0.48$ & $0.25$ & $0.00$ & - & $0.9261$ & $0.54$ & $0.41$ & $0.02$ \\
        & Conf. Sampling & - & - &  $0.9951$ & $0.94$ & $0.33$ & $0.03$ & - & $0.9950$ & $0.96$ & $0.14$ & $0.00$\\
        & Non-Ex. CS & IP &  $3.93$ & $0.8251$ & $0.16$ & $0.63$ & $0.26$  & $11.90$ & $0.8815$ & $0.24$ & $0.67$ & $0.03$ \\
         & & $l_2$ & $512.14$ & $0.8334$ & $0.17$ & $0.60$ & $0.06$ & $419.91$ & $0.8468$ & $0.18$ & $0.61$ & $0.05$ \\
        & & cos & $2.54$ & $0.8371$ & $0.17$ & $0.63$ & $0.06$ & $3.53$ & $0.8540$ & $0.17$ & $0.62$ & $0.04$ \\[0.1cm]
        \midrule
        \multirow{5}{*}{\rotatebox[origin=c]{90}{M2M100$_\text{(1.2B)}$}} & Nucleus Sampling & - & - & $ 0.8339$ & $0.38$ & $0.00$ &  $0.08$ & - & $0.7962$ & $0.42$ & $0.03$ & $0.10$ \\
        & Conf. Sampling & - & - & $0.9993$ & $0.99$ & $0.34$ & $0.00$ & - & $0.9998$ & $0.99$ & $0.60$ & $0.00$\\
        & Non-Ex. CS & IP & $15.79$ & $0.8861$ & $0.25$ & $0.71$ & $0.03$  & $10.45$ & $0.9129$ & $0.38$ & $0.72$ & $0.00$\\
        & & $l_2$  & $1123.45$ & $0.8874$ & $0.25$ & $0.72$ & $0.03$  & $605.97$ & $0.8896$ & $0.30$ & $0.76$ & $0.01$ \\
        & & cos & $3.21$ & $0.8858$ & $0.25$ & $0.72$ &  $0.03$ & $1.48$ & $0.8897$ & $0.30$ & $0.75$ & $0.01$ \\
        \bottomrule[0.05cm]
    \end{tabular}%
    }
    \caption{Coverage results for the de $\rightarrow$ en and ja $\rightarrow$ en MT tasks. We report the best found temperature $\tau$ while keeping the confidence level $\alpha$ and number of neighbors $k=100$ fixed. We also show the coverage percentage along with the avg. prediction set size as a proportion of the entire vocabulary ($\varnothing$ \textsc{Width}) as well as ECG and SSC. Tested distance metrics are inner product (IP), (squared) $l_2$ distance, and cosine similarity (cos).}\label{tab:coverage-results-mt}
\end{table*}

\begin{table}[htb!]
    \centering
    \resizebox{0.475\textwidth}{!}{
    \renewcommand{\arraystretch}{1.4}
        \begin{tabular}{@{}lllrcccc@{}}
        \toprule[0.05cm]
         & & & \multicolumn{5}{c}{\textsc{OpenWebText}} \\
         \cmidrule(lr){4-8}
         & Method & Dist. & $\tau$ & \% \textsc{Cov.} & $\varnothing$ \textsc{Width} $\downarrow$ &  \textsc{Scc} $\uparrow$ & \textsc{Ecg} $\downarrow$ \\
        \midrule 
        \multirow{5}{*}{\rotatebox[origin=c]{90}{OPT$_\text{(350M)}$}} & Nucl. Sampl. & - & - & $0.8913$ & $0.05$ & $0.71$ & $0.01$ \\
        & Conf. Sampl. & - & - & $0.9913$ & $0.90$ & $0.91$ &  $0.00$ \\
        & Non-Ex. CS & IP & $4.99$ & $0.9352$ & $0.19$ & $0.80$ & $0.0$ \\
        & & $l_2$ & $0.31 \times 10^4$ & $0.9425$ & $0.17$ & $0.80$ & $0.0$  \\ 
        & & cos & $4.98$ & $0.9370$ & $0.15$ & $0.83$& $0.0$ \\
        \midrule
        \multirow{5}{*}{\rotatebox[origin=c]{90}{OPT$_\text{(1.3B)}$}} & Nucl. Sampl. & - & - & $0.8952$ & $0.05$ & $0.00$ & $0.01$ \\
        & Conf. Sampl. & - & - & $0.9905$ & $0.88$ & $0.95$ &  $0.0$ \\
        & Non-Ex. CS  & IP & $0.48$ & $0.9689$ & $0.59$ & $0.84$ & $0.0$  \\
        & & $l_2$ & $1.55 \times 10^4$ & $0.9539$ & $0.20$ & $0.83$ & $0.0$  \\
        & & cos & $0.11$ & $0.9512$ & $0.20$ & $0.875$ & $0.0$ \\
        \bottomrule[0.05cm]
    \end{tabular}%
    }
    \caption{Coverage results for the LM task. We report the best found temperature $\tau$ while keeping the confidence level $\alpha$ and number of neighbors $k=100$ fixed. We also show the coverage percentage along with the avg. prediction set size as a proportion of the entire vocabulary ($\varnothing$ \textsc{Width}) as well as the ECG and SSC metrics. Tested distance metrics are inner product (IP), (squared) $l_2$ distance and cos. similarity (cos).}
    \label{tab:coverage-results-lm}
\end{table}

First of all, we demonstrate that the retrieved information from the data store enables us to successfully apply the proposed method. 
\emph{Coverage} is an important notion in conformal prediction, referring to the correct label being covered by a prediction set or intervals. 
Since we can always achieve trivial coverage by choosing the largest possible prediction set, an ideal method would strike a balance between high coverage and small prediction sets.
While it is not possible to measure coverage in a free generation setting (see next section), we can assess whether the correct class is contained in the prediction set if we feed the actual reference tokens into the decoder and check whether we include the true continuation.\footnote{We emphasize that access to gold tokens is not required by our method and only done here to measure the actual coverage.}
For our MT task, this is reminiscent of an interactive translation prediction setup \citep{knowles2016neural, peris2017interactive, knowles2019user}, where we would like to suggest possible continuations to a translator, suggesting the next word from a set of words that (a) contains plausible options and (b) is limited in size, in order to restrict the complexity for the end user.
Before we run our experiments, we need to determine $\tau$, which we tune on the calibration set using a stochastic hill-climbing procedure described in \cref{app:temperature-search}.  
We compare our \emph{non-exchangeable conformal nucleus sampling} (\emph{Non-Ex. CS}) with nucleus sampling \citep{holtzman2020curious} and conformal nucleus sampling (\emph{Conf. Sampl.};~\citealp{ravfogel2023conformal}). The latter bin predictions on a calibration set by the entropy of the output distribution, and compute one $\hat{q}$ per such entropy bin using the standard conformal procedure given in the beginning of \cref{sec:background}. 
\looseness=-1

\begin{figure*}[htb!]
    \centering

    \begin{subfigure}[t]{\columnwidth}
        \centering
        \includegraphics[width=0.9\columnwidth]{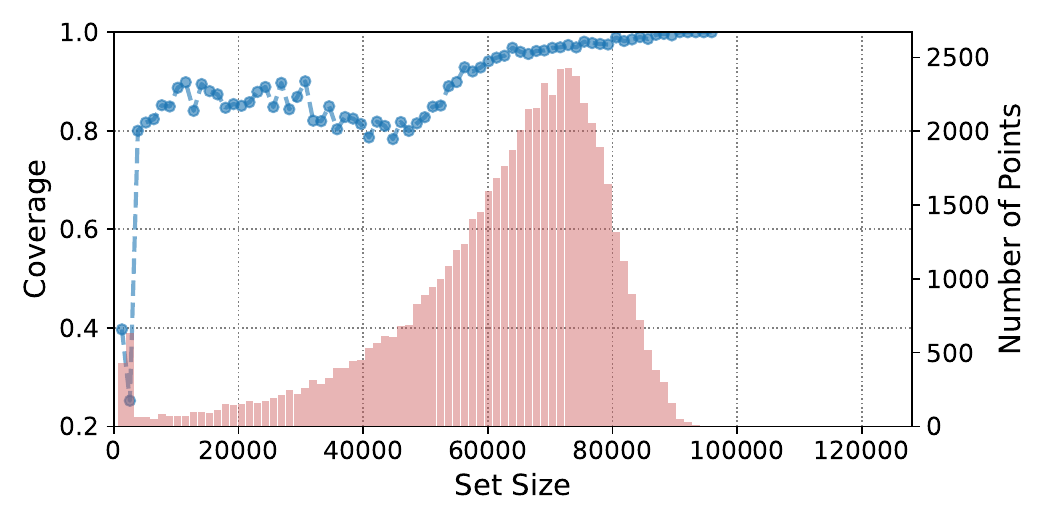}
        \caption{Nucleus Sampling on de $\rightarrow$ en.}
        \label{subfig:stratified-coverage-deen-nucleus}
    \end{subfigure}
    \hfill
    \begin{subfigure}[t]{\columnwidth}
        \centering
        \includegraphics[width=0.9\columnwidth]{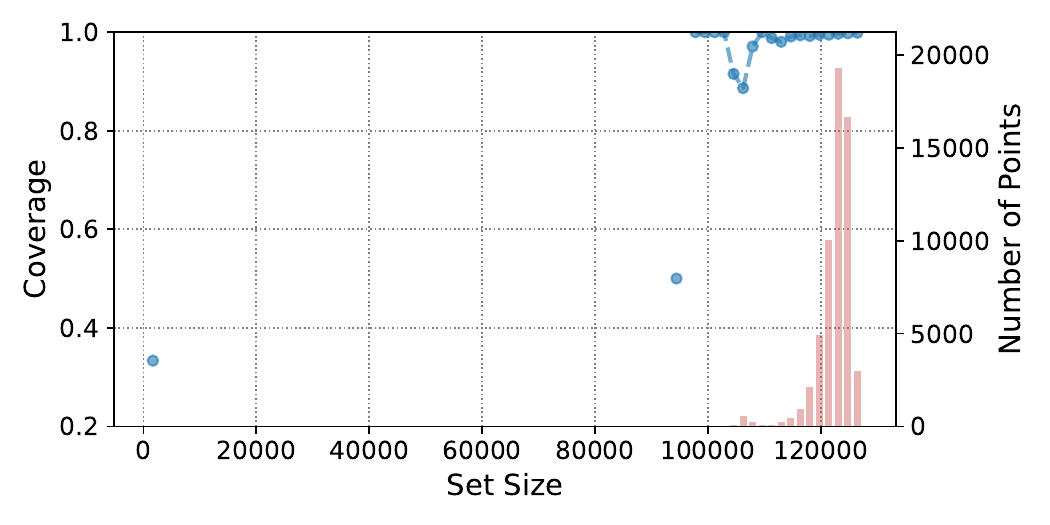}
        \caption{Conformal Nucleus Sampling on de $\rightarrow$ en.}
        \label{subfig:stratified-coverage-deen-conformal-nucleus}
    \end{subfigure}

    \begin{subfigure}[t]{\columnwidth}
        \centering
        \includegraphics[width=0.9\columnwidth]{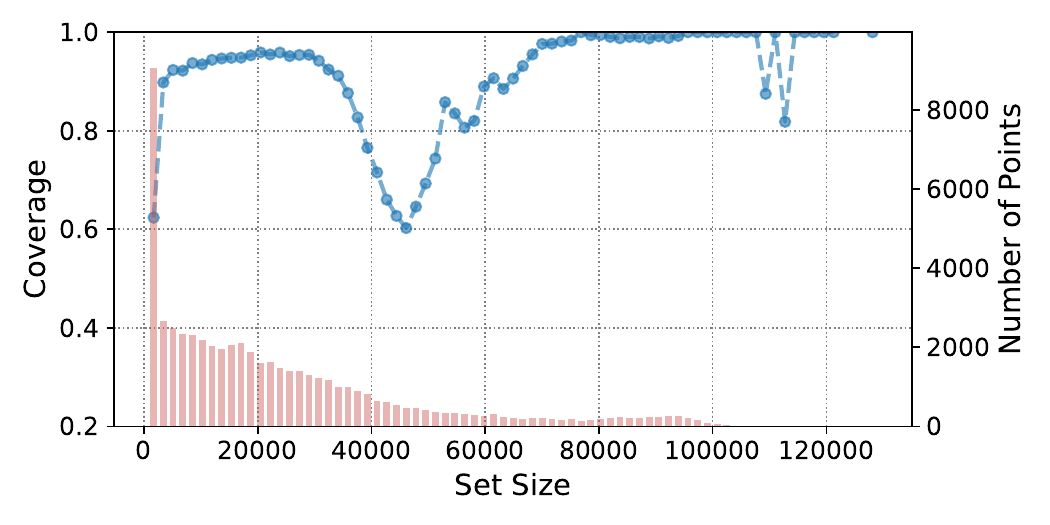}
        \caption{Non-Ex. Conformal Sampling on de $\rightarrow$ en.}
        \label{subfig:stratified-coverage-deen}
    \end{subfigure}
    \hfill
    \begin{subfigure}[t]{\columnwidth}
        \centering
        \includegraphics[width=0.9\columnwidth]{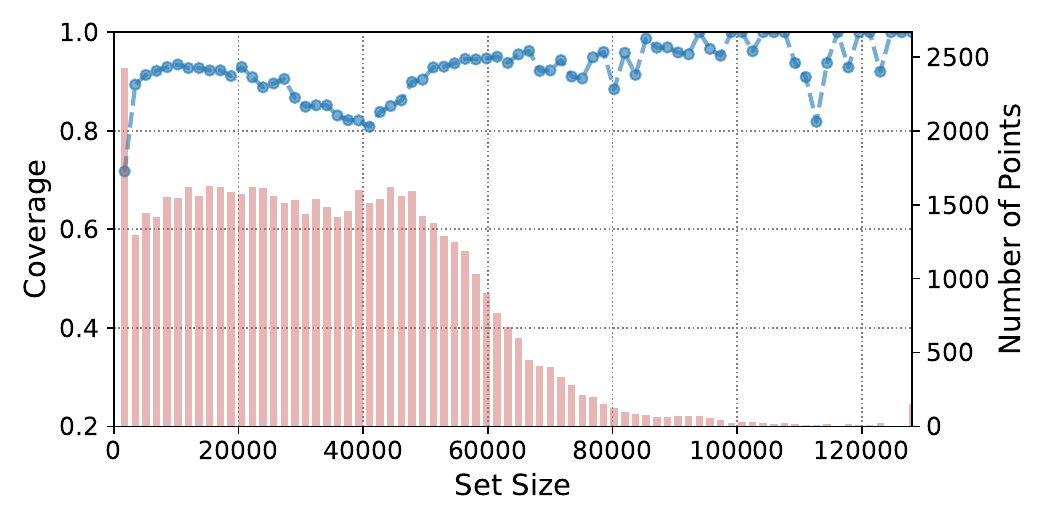}
        \caption{Non-Ex. CS on de $\rightarrow$ en with M2M100$_\text{(1.2B)}$.}
        \label{subfig:stratified-coverage-deen-large}
    \end{subfigure}
    \caption{Conditional coverage for the M2M100 on de $\rightarrow$ en with the small 418M model (\cref{subfig:stratified-coverage-deen-nucleus,subfig:stratified-coverage-deen-conformal-nucleus,subfig:stratified-coverage-deen}) and using the bigger 1.2B model (\cref{subfig:stratified-coverage-deen-large}). We aggregate predictions by set size using $75$ equally-spaced bins in total. The blue curve shows the conditional coverage per bin, whereas red bars show the number of binned predictions. }\label{fig:conditional-coverage}
\end{figure*}

\paragraph{Evaluation.} We measure the total coverage using different distance metrics, namely, squared $l_2$ distance, normalized inner product, and cosine similarity (see \cref{tab:coverage-results-mt,tab:coverage-results-lm}),\footnote{For inner product and cosine similarity, we follow the same form as \cref{eq:weight-equation}, omitting the minus. We normalize the inner product by the square root of the latent dimension.} as well as binning predictions by set size and then measuring the per-bin coverage in \cref{fig:conditional-coverage} (more results given in \cref{app:coverage-experiments}).
We also summarize the plots in \cref{fig:conditional-coverage} via the \emph{Expected Coverage Gap} (ECG)\footnote{This is inspired by the expected calibration error \citep{guo2017calibration}, comparing coverage to $1 - \alpha$, where overcoverage is not penalized due to \cref{eq:conformal-prediction-guarantees}'s lower bound.} that we define as 

\begin{equation}\label{eq:ecg}
    \text{ECG} = \sum_{b=1}^B \frac{|\mathcal{B}_b|}{N} \max \Big( 1 - \alpha - \text{Coverage}\big(\mathcal{B}_b\big), 0 \Big),
\end{equation}
\noindent where $\mathcal{B}_b$ denotes a single bin and $N$ the total number of considered predictions in the dataset.\footnote{Since conformal prediction produces a \emph{lower} bound on the coverage, we do not include overcoverage in \cref{eq:ecg}.} 
The ECG thus captures the average weighted amount of undercoverage across bins.
In our experiments, we use $75$ bins in total. 
The same bins are used to also evaluate the \emph{Size-Stratified Coverage metric} (SSC) proposed by \citet{angelopoulos2021uncertainty}, with a well-calibrated method resulting in a SCC close to the desired coverage $1-\alpha$:

\begin{equation}
    \text{SCC} = \min_{b \in \{1,\ldots, B\}} \text{Coverage}\big(\mathcal{B}_b\big).
\end{equation}

\noindent We can therefore understand the SCC as the worst-case coverage across all considered bins.
We present some additional experiments where we assess the impact of key hyperparameters in \cref{app:ablations}.

\paragraph{Results.} We found our method to miss the desired coverage of $90 \%$ for MT by $8 \%$ or less. 
Beyond the reported values, we were not able to further increase coverage by varying the temperature parameter without avoiding trivial coverage (i.e.\@, defaulting to very large set sizes), which is likely due to the impossible-to-estimate coverage in \cref{eq:non-exchangeable-conformal-prediction-guarantees}.
Most notably, our method was able to achieve better SCC scores while maintaining considerably smaller prediction sets than the baselines on average. 
The reason for this is illustrated in \cref{fig:conditional-coverage}:
while standard nucleus sampling produces some prediction sets that are small, the total coverage seems to mostly be achieved by creating prediction sets between 60k--80k tokens. 
The behavior of conformal nucleus sampling by \citet{ravfogel2023conformal} is even more extreme in this regard, while our method focuses on producing smaller prediction sets, with the frequency of larger set sizes decreasing gracefully. 
In \cref{subfig:stratified-coverage-deen-large}, we can see that the larger M2M100 models also tend to produce larger prediction sets, but still noticeably smaller than the baselines. Importantly, for both M2M100 models, even very small prediction sets (size $\leq 1000$) achieve non-trivial coverage, unlike the baseline methods.
For LM, we always found the model to slightly \emph{over}cover.
This does not contradict the desired lower bound on the coverage in \cref{eq:non-exchangeable-conformal-prediction-guarantees} and suggests a more negligible distributional drift.
While nucleus sampling produces the smallest average prediction sets, we can see that based on the SCC values some strata remain undercovered. 
Instead, our method is able to strike a balance between stratified coverage and prediction set size.
With respect to distance measures, we find that the difference between them is minimal, indicating that the quality largely depends on the retrieved local neighborhood of the decoder encoding and that finding the right temperature can help to tune the models to approximate the desired coverage.
We would now like to find out whether this neighborhood retrieval mechanism can prove to be robust under distributional shift as well.
Since we did not observe notable differences between the distance metrics, we continue with the $l_2$ distance.

\subsection{Coverage Under Shift}\label{sec:shift-robustness}

\begin{figure*}
    \centering
    \begin{subfigure}[t]{0.9\textwidth}
        \centering
        \includegraphics[trim={0 1cm 0 0},clip,width=\textwidth]{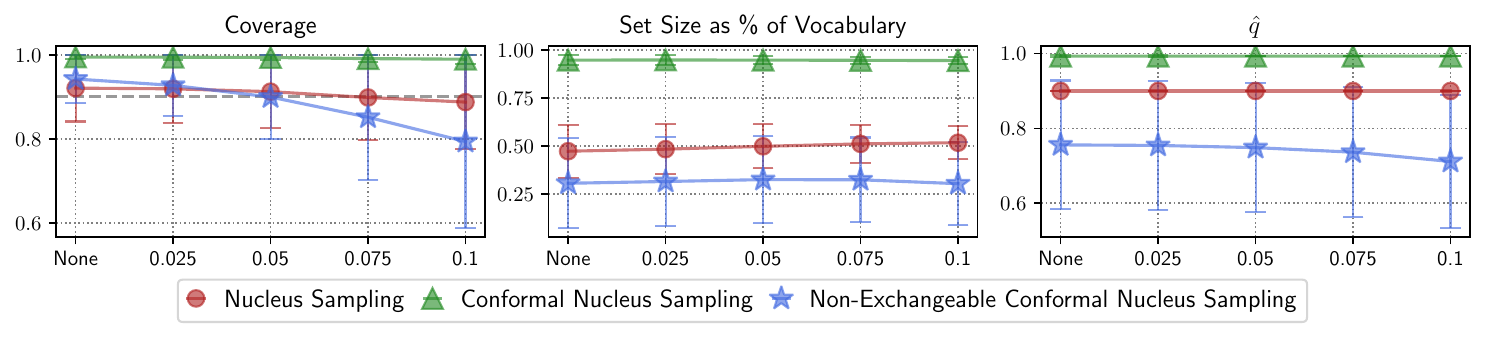}
    \end{subfigure}
    \begin{subfigure}[t]{\textwidth}
        \centering
        \includegraphics[width=0.9\textwidth]{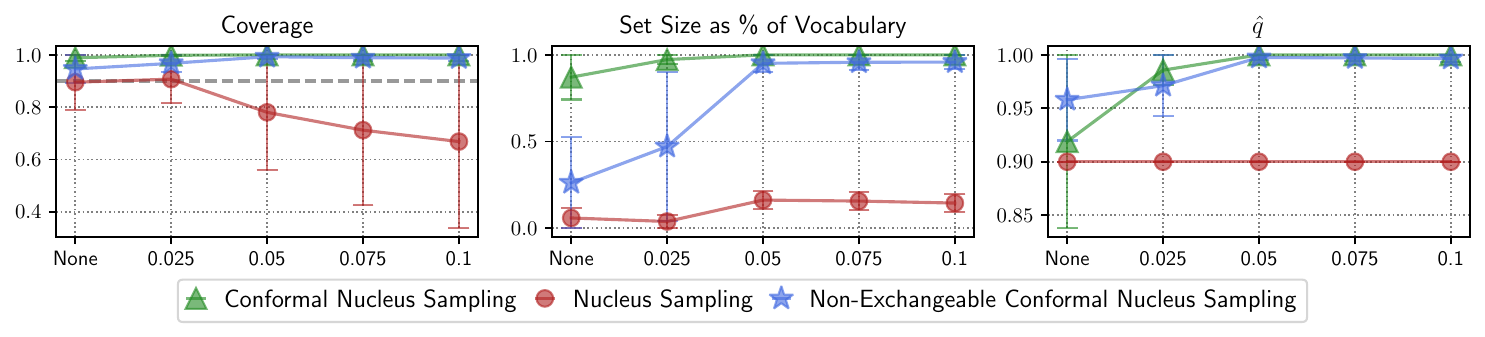}
    \end{subfigure}
    \caption{Coverage, average set size and $\hat{q}$ based on the noise level on the de $\rightarrow$ en MT task (top) and open text generation task (bottom). Error bars show one standard deviation.}\label{fig:noise-experiment-deen}
\end{figure*}

To demonstrate how the retrieval of nearest neighbors can help to maintain coverage under distributional shift, we add Gaussian noise of increasing variance---and therefore intensity---to the last decoder hidden embeddings (for MT) and the input embeddings (LM).\footnote{A similar approach can be found for instance in the work of \citet{hahn2019self, zhang2023text} or by \citet{ovadia2019can,hendrycks2019benchmarking} in a computer vision context.}
This way, we are able to simulate distributional drift while still keeping the original sequence of input tokens intact, allowing us to measure the actual coverage.
We show the achieved coverage along with the average set size (as a percentage of the total vocabulary) and the average quantile $\hat{q}$ in \cref{fig:noise-experiment-deen}.
We can see that the conformal sampling method deteriorates into returning the full vocabulary as a prediction set. Thus it behaves similarly to simple sampling as indicated by the $\hat{q}$ values being close to $1$. Nucleus sampling provides smaller prediction sets compared to conformal sampling, but they seem invariant to noise. As such, the method is not robust to noise injection in the open text generation task, and the obtained coverage deteriorates with noise variance $\geq 0.025$. Instead, the use of nearest neighbors allows for the estimation of prediction sets that are small but amenable to increase, such that the obtained coverage remains close to the desired one. We can specifically observe that the prediction set size increases considerably to mitigate the injected noise in the open-text generation case.  

\begin{table}
    \centering
    \resizebox{0.995\columnwidth}{!}{
    \renewcommand{\arraystretch}{1.4}
    \begin{tabular}{@{}lcccccc@{}}
        \toprule[0.05cm]
         & \multicolumn{5}{c}{\textsc{Noise level }} \\
         \cmidrule(lr){2-6}
          & \textsc{None} & $0.025$ & $0.05$ & $0.075$ & $0.1$ \\
        \midrule 
        $\varnothing$ Entropy & $8.46$ & $8.71$ & $9.20$ & $9.71$ & $10.08$ \\
        \midrule[0.005cm]
        Nucl. Sampl. $(\rho)$  & $0.87$ & $0.86$ & $0.84$ & $0.82$ & $0.81$ \\
        Conf. Sampl. $(\rho)$ & $0.60$ & $0.60$ & $0.60$ & $0.57$ & $0.55$\\
        Non-Ex. CS $(\rho)$ & $-0.14$ & $-0.18$ & $-0.27$ & $-0.37$ & $-0.45$ \\
        \bottomrule
    \end{tabular}
    }
    \caption{Average entropy of 400M M2M100 model on de $\rightarrow$ en per noise level as well as the Spearman's $\rho$ correlation coefficients between the predictive entropy and the prediction set size of the different methods. All results are significant with $p < 0.0001$.}\label{tab:analysis}
\end{table}

\paragraph{Neighbor Retrieval.} We further analyze how the retrieval enables this flexibility by relating it to the entropy of the output distribution of the 400M parameters M2M100 on German to English. 
Intuitively, the baseline methods, faced by high-entropy output distributions, need to produce wide prediction sets in order to maintain coverage. 
In fact, we report such results by correlating entropy levels and prediction set sizes using Spearman's $\rho$ in \cref{tab:analysis}, showing strong positive correlations.
Our method in contrast shows consistently an \emph{anti}correlation between these two quantities, enabled by decoupling the creation of prediction sets from statistics of the output distribution to instead considering the non-conformity scores of similar subsequences.
The fact that the prediction set size is not just dependent on the entropy of the predictions while maintaining coverage demonstrates the value of the nearest neighbors:
In this way, model uncertainty becomes more flexible and is corroborated by evidence gained from similar inputs.
 
\subsection{Generation Quality}\label{sec:generation-quality}

\begin{table*}[htb]
    \centering

    \begin{subfigure}[t]{0.61\textwidth}
        \resizebox{0.995\textwidth}{!}{
        \renewcommand{\arraystretch}{1.4}
        \begin{tabular}{@{}llcccccc@{}}
            \toprule[0.05cm]
             & &  \multicolumn{3}{c}{de $ \rightarrow $ en} & \multicolumn{3}{c}{ja $ \rightarrow $ en} \\
             \cmidrule(lr){3-5}\cmidrule(lr){6-8} 
             & Method & \textsc{Bleu} $\uparrow$ & \textsc{Comet} $\uparrow$ & \textsc{ChrF} $\uparrow$ & \textsc{Bleu} $\uparrow$ & \textsc{Comet} $\uparrow$ & \textsc{ChrF} $\uparrow$ \\
            \midrule 
            \multirow{7}{*}{\rotatebox[origin=c]{90}{M2M100$_\text{(400m)}$}} & Beam search & $28.53$ & $0.88$ & $ 55.58$ & $11.37$ & $0.63$ & $37.74$ \\
             & Greedy & $27.81$ & $0.9$ & $54.9$ & $10.73$ & $0.58$ & $36.5$ \\[0.2cm]
             & Nucleus Sampling & $27.63 {\scriptstyle\ \pm 0.03}$ & $0.89 {\scriptstyle\ \pm 0.01}$ & $54.80 {\scriptstyle\ \pm 0.07}$ & $10.61 {\scriptstyle\ \pm 0.15}$ & $0.59 {\scriptstyle\ \pm 0.01}$ & $36.52{\scriptstyle\ \pm 0.19}$ \\ 
             & Top-$k$ Sampling & $27.63 {\scriptstyle\ \pm 0.03}$ & $0.89 {\scriptstyle\ \pm 0.01}$ & $54.79 {\scriptstyle\ \pm 0.07}$ & $10.61 {\scriptstyle\ \pm 0.15}$ & $0.59 {\scriptstyle\ \pm 0.01}$ & $36.52{\scriptstyle\ \pm 0.19}$ \\ 
             & Conf. Sampling &  $27.63 {\scriptstyle\ \pm 0.03}$ & $0.89 {\scriptstyle\ \pm 0.01}$ & $54.80 {\scriptstyle\ \pm 0.07}$ & $10.61 {\scriptstyle\ \pm 0.15}$ & $0.59 {\scriptstyle\ \pm 0.01}$ & $36.52 {\scriptstyle\ \pm 0.19}$\\ 
             & Const. Weight CS$^*$ & $27.63 {\scriptstyle\ \pm 0.03}$ & $0.89 {\scriptstyle\ \pm 0.01}$ & $54.80 {\scriptstyle\ \pm 0.07}$ & $10.61 {\scriptstyle\ \pm 0.15}$ & $0.59 {\scriptstyle\ \pm 0.01}$ &  $36.52 {\scriptstyle\ \pm 0.19}$ \\
             & Non-Ex. CS$^*$ & $27.65 {\scriptstyle\ \pm 0.10}$ & $0.90 {\scriptstyle\ \pm 0. 
 01}$ & $54.82 {\scriptstyle\ \pm 0.14}$ & $\underline{10.74} {\scriptstyle\ \pm 0.11}$ & $0.59 {\scriptstyle\ \pm 0.01}$ & $36.61 {\scriptstyle\ \pm 0.08}$ \\ 
             \midrule
             \multirow{7}{*}{\rotatebox[origin=c]{90}{M2M100$_\text{(1.2B)}$}} & Beam search & $30.89$ & $0.9$ & $56.8$ & $13.76$ & $0.63$ & $40.43$ \\
             & Greedy & $29.52$ & $0.9$ & $55.67$ & $12.94$ & $0.6$ & $39.91$  \\[0.2cm]
             & Nucleus Sampling & $29.37{\scriptstyle\  \pm 0.12}$ &  $0.90{\scriptstyle\  \pm 0.00}$ & $55.55{\scriptstyle\  \pm 0.11}$ & $10.61{\scriptstyle\  \pm 0.15}$ & $0.59{\scriptstyle\  \pm 0.01}$ & $36.52{\scriptstyle\  \pm 0.19}$ \\
             & Top-$k$ Sampling & $29.53{\scriptstyle\  \pm 0.00}$ & $0.90{\scriptstyle\  \pm 0.00}$ & $55.67{\scriptstyle\  \pm 0.00}$ & $12.91{\scriptstyle\  \pm 0.08}$ & $0.60{\scriptstyle\  \pm 0.01}$ & $39.95{\scriptstyle\  \pm 0.00}$ \\
             & Conf. Sampling & $29.37{\scriptstyle\  \pm 0.12}$ & $0.90{\scriptstyle\ \pm 0.00}$ & $55.55{\scriptstyle\  \pm 0.11}$ & $12.91{\scriptstyle\  \pm 0.08}$ & $0.60{\scriptstyle\  \pm 0.00}$ & $39.95{\scriptstyle\  \pm 0.08}$\\
             & Const. Weight CS$^*$ & $29.37{\scriptstyle\  \pm 0.12}$ & $0.90{\scriptstyle\ \pm 0.00}$ & $55.55{\scriptstyle\ \pm 0.11}$ & $12.91{\scriptstyle\ \pm 0.08}$ & $0.60 {\scriptstyle\ \pm 0.01}$ & $39.95 {\scriptstyle\ \pm 0.08}$ \\
             & Non-Ex. CS$^*$ & $ 29.37 {\scriptstyle\ \pm 0.12}$ & $0.90 {\scriptstyle\ \pm 0.00}$ & $55.55 {\scriptstyle\ \pm 0.11}$ & $12.91 {\scriptstyle\ \pm 0.08}$ & $0.60 {\scriptstyle\ \pm 0.01}$ & $39.95 {\scriptstyle\ \pm 0.08}$ \\ 
            \bottomrule[0.05cm]
        \end{tabular}%
        }
        \caption{Generation results for the de $\rightarrow$ en and ja $\rightarrow$ en translation tasks.}\label{tab:generation-results-mt}
    \end{subfigure}
    \hfill
    \begin{subfigure}[t]{0.36\textwidth}
        \centering
        \resizebox{0.995\textwidth}{!}{
        \renewcommand{\arraystretch}{1.4}
        \begin{tabular}{@{}llccc@{}}
            \toprule[0.05cm]
             & & \multicolumn{3}{c}{\textsc{OpenWebText}} \\
             \cmidrule(lr){3-5}
             & Method & \textsc{MAUVE} $\uparrow$  & \textsc{BERTscore} $F_1$ $\uparrow$  \\
            \midrule 
           \multirow{6}{*}{\rotatebox[origin=c]{90}{OPT$_\text{(350M)}$}} & 
           Beam search & $0.12$ & $0.79$ \\
           & Greedy & $0.02$ & $0.79$\\[0.2cm] 
             & Nucleus Sampling & $0.91 {\scriptstyle\ \pm 0.02}$ & $0.80{\scriptstyle\ \pm 0.00}$ \\ 
             & Top-$k$ Sampling & $0.90 {\scriptstyle\ \pm 0.03}$ & $\underline{0.80} {\scriptstyle\ \pm 0.00}$ \\ 
             & Conf. Sampling & $0.91 {\scriptstyle\ \pm 0.02}$ & $0.80 {\scriptstyle\ \pm 0.00}$ \\ 
             & Const. Weight CS$^*$  & $0.91 {\scriptstyle\ \pm 0.02}$ & $0.80 {\scriptstyle\ \pm 0.00}$\\ 
             & Non-Ex. CS$^*$ & $0.92 {\scriptstyle\ \pm 0.01}$ & $0.80 {\scriptstyle\ \pm 0.00}$  \\ 
            \midrule
            \multirow{6}{*}{\rotatebox[origin=c]{90}{OPT$_\text{(1.3B)}$}} & Beam search & $0.17$ & $0.80$\\
            & Greedy & $0.05$ & $0.79$ \\[0.2cm] 
             & Nucleus Sampling & $0.91 {\scriptstyle\ \pm 0.02}$ & $0.80 {\scriptstyle\ \pm 0.00}$\\ 
             & Top-$k$ Sampling & $0.93 {\scriptstyle\ \pm 0.01}$ & $\underline{0.81} {\scriptstyle\ \pm 0.00}$\\ 
             & Conf. Sampling & $0.93 {\scriptstyle\ \pm 0.01}$ & $0.80 {\scriptstyle\ \pm 0.00}$ \\ 
             & Const. Weight CS$^*$ & $0.91 {\scriptstyle\ \pm 0.02}$ & $0.80 {\scriptstyle\ \pm 0.00}$ \\ 
             & Non-Ex. CS$^*$ & $0.92 {\scriptstyle\ \pm 0.01}$ & $0.81 {\scriptstyle\ \pm 0.00}$\\ 
              \bottomrule[0.05cm]
        \end{tabular}%
        }
        \caption{Results for the open text generation.}
        \label{tab:generation-results-lm}
    \end{subfigure}
    \caption{Generation results for the two tasks. We report performance using $5$ beams for beam-search, top-$k$ sampling with $k=10$, and nucleus sampling with $p=0.9$. Conformal methods all use $\alpha =  0.1$, with non-exchangeable variants retrieving $100$ neighbors. MT results for sampling use a softmax temperature of $0.1$. Our  methods are marked with $^*$. Results using $5$ different seeds that are stat. significant according to the ASO test \citep{del2018optimal, dror2019deep, ulmer2022deep} with a confidence level of $0.95$ and threshold $\varepsilon_\text{min} \le 0.3$ are underlined.}
    \label{tab:translations-results}
\end{table*}

Crucially, our method should not degrade and potentially even improve generation quality. Thus, we evaluate generation quality for the same tasks without supplying the gold prefix. For language modeling, we follow \citet{ravfogel2023conformal} and use the first $35$ tokens from the original sentence as input.
We compare against a set of generation strategies including top-$k$ sampling \citep{fan2018hierarchical, holtzmann2018learning, radford2019language}, nucleus sampling and conformal nucleus sampling. We also test a variant of our method using constant weights $w_k = 1$ for retrieved neighbors (\emph{Const. Weight CS}) to assess the impact of the weighted neighbor retrieval 
procedure.
We further compare with beam search \citep{medress1977speech, graves2012sequence} with a softmax temperature of $0.1$, and greedy decoding.
Evaluation is performed using BLEU \citep{papineni2002bleu}, COMET-22 \citep{rei2020comet, rei2022comet22} and chrF \citep{popovic-2017-chrf} for MT as well as MAUVE \citep{pillutla-etal:mauve:neurips2021} and BERTscore \citep{zhang2020bertscore} for text generation.\footnote{All metrics except for COMET were used through Hugging Face \texttt{evaluate}. MAUVE uses \texttt{gpt2} as a featurizer.}

\paragraph{Results.} We show the results for the different methods in \cref{tab:translations-results}.
We see that beam search outperforms all sampling methods for MT. 
This corroborates previous work by \citet{shaham2022you} who argue that (nucleus) sampling methods, by pruning only the bottom percentile of the
token distribution, introduce some degree of randomness that is beneficial for open text generation but may be less optimal for conditional language generation, where the desired output is constrained and exact matching generations are preferred (which is the case for MT).
Among sampling methods, we find nucleus sampling and conformal sampling to perform similarly (being in agreement with the findings of \citealp{ravfogel2023conformal}) but are sometimes on par or even outperformed by our non-exchangeable conformal sampling for MT.
For text generation, our method performs best for the smaller OPT model but is slightly beaten by conformal nucleus sampling in terms of MAUVE.
When using constant weights, performance deteriorates to the conformal sampling setup, emphasizing the importance of not considering all conformity scores equally when computing $\hat{q}$, even though the effect seems to be less pronounced for larger models.
This illustrates the benefit of creating flexible prediction sets that are adapted on token-basis, suggesting that both the latent space neighborhoods as well as the conformity scores are informative.
We discuss examples of generated text in \cref{app:qualitative-analysis}.

\section{Discussion}\label{sec:discussion}

Our experiments have shown that despite the absence of i.i.d.\@ data in NLG and the loss in coverage induced by using dynamic calibration sets, the resulting coverage is still close to the pre-specified desired level for both LM and MT.
Additionally, even though the coverage gap predicted by the method of \citet{barber2022conformal} is infeasible to compute for us, we did not observe any critical degradation in practice. 
Further, we demonstrated how sampling from these calibrated prediction sets performs similarly or better than other sampling methods.
Even though our method is still outperformed by beam search in the MT setting, previous work such as minimum Bayes risk (MBR) decoding has shown how multiple samples can be re-ranked to produce better outputs \citep{
kumar2004minimum, eikema2020is, freitag2023epsilon, fernandes2022quality}.
Additionally, recent dialogue systems based on LLMs use sampling instead of beam search for generation.
Since our prediction sets are more flexible and generally tighter, our results serve as a starting point for future work. 
For instance, our technique could be used with non-conformity scores that do not consider token probabilities alone (e.g.\@ \citealp{meister2023locally}) or using prediction set widths as a proxy for uncertainty \citep{angelopoulos2020uncertainty}.

\section{Conclusion}\label{sec:conclusion}

We successfully demonstrated the application of a non-exchangeable variant of conformal prediction to machine translation and language modeling with the help of $k$-NN retrieval.
We showed our method to be able to maintain the desired coverage best across different dataset strata while keeping prediction sets smaller than other sampling methods, all while providing theoretical coverage guarantees about coverage that other comparable methods lack.
We validated our method to produce encouraging results for generation tasks.
Lastly, we analyzed the behavior under distributional drift, showing how the $k$-NN retrieval maintains desirable properties for the estimated prediction sets.
We see our method as a step to provide a more principled way to perform sampling with conformal guarantees under more realistic assumptions.

\section*{Limitations}

We highlight two main limitations of our work here:
Potential issues arising from different kinds of dataset shift as well as efficiency concerns.

\paragraph{Distributional Drifts.} Even though any loss of coverage due to the term quantifying distributional drift in \cref{eq:non-exchangeable-conformal-prediction-guarantees} was limited in our experiments (see \cref{sec:retrieval-quality,sec:shift-robustness}), this might not hold across all possible setups. 
As long as we cannot feasibly approximate the shift penalty, it is impossible to determine a priori whether the loss of coverage might prove to be detrimental, and would have to be checked in a similar way as in our experiments. 
Furthermore, we only consider shifts between the models' training distributions and test data distributions here, while many other, unconsidered kinds of shifts exist \citep{moreno2012unifying, hupkes2022state}.

\paragraph{Computational Efficiency.} Even using optimized tools such as FAISS \citep{johnson2019billion}, moving the conformal prediction calibration step to inference incurs additional computational cost during generation.
Nevertheless, works such as \citet{he2021efficient, henrique2022chunk} show that there are several ways to improve the efficiency of $k$-NN approaches, and we leave such explorations to future work.

\section*{Ethical Considerations}

The main promise of conformal prediction lies in its correctness---i.e.\@ producing prediction sets that contain the correct prediction and are thus reliable.
In an application, this could potentially create a false sense of security. 
On the one hand, the conformal guarantee holds in expectation, and not necessarily on a per-sample basis. 
On the other hand, our experiments have demonstrated that coverage might also not hold when distributional shifts are at work or when looking at specific subpopulations.
Therefore, any application should certify that coverage is maintained for potentially sensitive inputs.

\section*{Acknowledgements}

We thank the anonymous reviewers for the constructive feedback and useful discussions.
This work was supported by EU’s Horizon Europe Research and Innovation Actions (UTTER, contract 101070631), by the project DECOLLAGE (ERC-2022-CoG 101088763), by the Portuguese Recovery and Resilience Plan through project C645008882-00000055 (Center for Responsible AI), and by Fundac\~{a}o para a Ci\^{e}ncia e Tecnologia through contract UIDB/50008/2020.

\bibliography{anthology,custom}

\appendix

\section{Appendix}

Aside from \cref{app:adaptive-prediction-sets} giving more detail on the construction of adaptive prediction sets, we use this appendix to bundle more details about experiments and their results.
\cref{app:temperature-search} details the procedure to determine the temperature in \cref{eq:weight-equation}.
We present more results from the experiments in \cref{sec:retrieval-quality} in \cref{app:coverage-experiments}.\\

We illustrate the overall algorithm in \cref{app:algorithm} and estimate environmental impact of our work in \cref{app:environmental-impact}.

\subsection{Adaptive Prediction Sets}\label{app:adaptive-prediction-sets}

Here we provide a more formal definition of the adaptive prediction sets.
Let $\pi$ be a permutation function mapping all possible output tokens $\{1, \ldots, C\}$ to the indices of a permuted version of the set, for which tokens are sorted by their probability under the model, descendingly. We define the non-conformity score as 
\begin{equation}\label{eq:adaptive-non-conformity-score}
        s_i = \sum_{j=1}^{\pi(y_t)} p_{\btheta}\big(\pi^{-1}(j)\big|\bx, y_{<t}\big).
\end{equation}
Since we only include the cumulative mass up until the gold label, the summation stops at $\pi(y)$. 
The prediction sets are then defined as 
\begin{align}\label{eq:adaptive-prediction-set}
    \mathcal{C}(\bx^*, y^*_{<t}) & = \Big\{\pi^{-1}(1), \ldots, \pi^{-1}(\hat{c})\Big\},
\end{align}
with $\hat{c} = \sup \{c^\prime\ |\ \sum_{j=1}^{c^\prime} p_{\btheta}(\pi^{-1}(j)\,|\,\bx^*, y^*_{<t}) < \hat{q} \} + 1$, where we add one extra class to avoid empty sets.

\subsection{Temperature Search}\label{app:temperature-search}

In order to determine the temperature used in \cref{eq:weight-equation} for the different distance metrics in \cref{tab:coverage-results-mt}, we adopt a variation of a simple hill-climbing procedure. 
Given user-defined bounds for the temperature search $\tau_\text{min}$ and $\tau_\text{max}$, we sample an initial candidate $\tau_0 \sim \mathcal{U}[\tau_\text{min}, \tau_\text{max}]$, and then evaluate the coverage of the method given the candidate on the first $100$ batches of the calibration dataset. 
The next candidate then is obtained via 

\begin{align}
    \tau_{t+1} & = \tau_{t} + \eta \cdot \varepsilon \cdot \text{sgn}\big(1 - \alpha - \text{Coverage}(\tau_t)\big); \nonumber \\
    \varepsilon & \sim \mathcal{N}(0, \tau_\text{max} - \tau_\text{min}), 
\end{align}

\noindent where $\eta$ is a predefined step size (in our case $0.1$) and $\text{Coverage}(\tau_t)$ the achieved coverage given a candidate $\tau_t$.
The final temperature is picked after a fixed number of steps ($t=20$ in our work) based on the smallest difference between achieved and desired coverage.\\

Overall, we found useful search ranges to differ greatly between datasets, models, and distance metrics, as illustrated by the reported values in \cref{tab:coverage-results-mt} and \cref{tab:coverage-results-lm}.
In general, the stochastic hill-climbing could also be replaced by a grid search, even though we sometimes found the best temperature to be ``hidden'' in a very specific value range.
It also has to be noted that temperature for the $l_2$ distance is the highest by far since FAISS returns \emph{squared} $l_2$ distances by default.

\subsection{Additional Coverage Results}\label{app:coverage-experiments}

\begin{figure}[htb!]
    \centering
    \begin{subfigure}[t]{\columnwidth}
        \centering
        \includegraphics[width=0.9975\columnwidth]{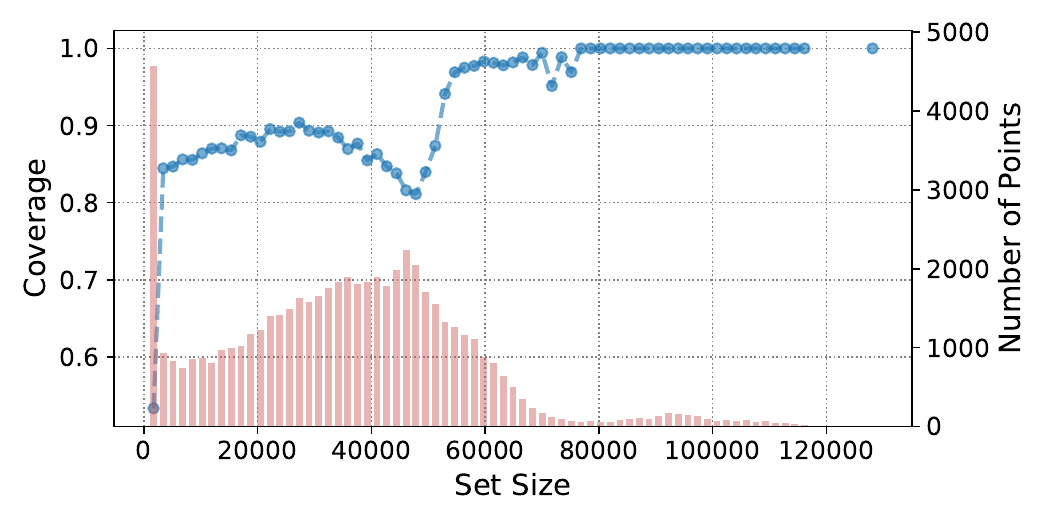}
        \caption{Conditional coverage of M2M100$_\text{(1.2B)}$ for de $\rightarrow$ en.}
        \label{subfig:stratified-coverage-deen-m2m00large}
    \end{subfigure}
    \begin{subfigure}[t]{\columnwidth}
        \centering
        \includegraphics[width=0.9975\columnwidth]{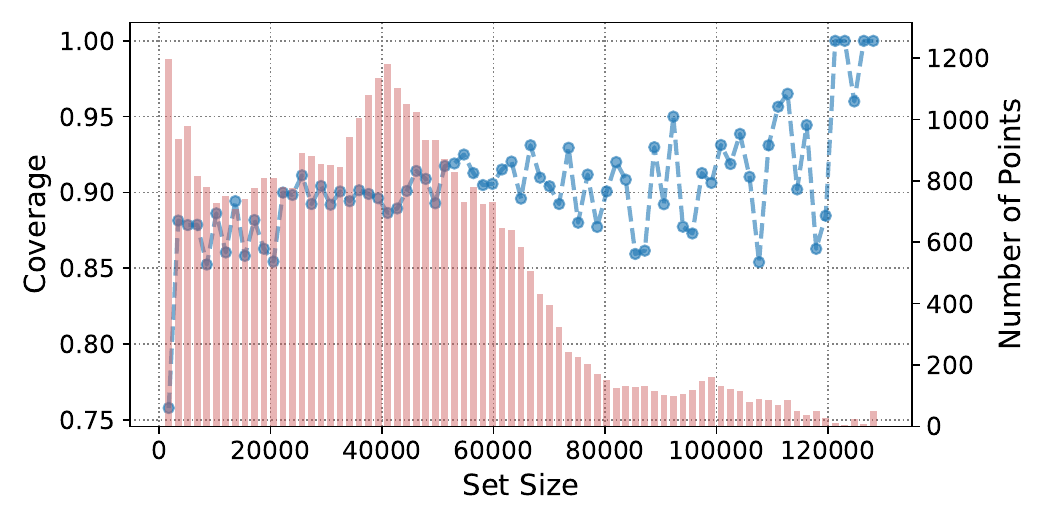}
        \caption{Conditional coverage of M2M100$_\text{(1.2B)}$ for ja $\rightarrow$ en.}
        \label{subfig:stratified-coverage-jaen-m2m00large}
    \end{subfigure}
    \begin{subfigure}[t]{\columnwidth}
        \centering
        \includegraphics[width=0.9975\columnwidth]{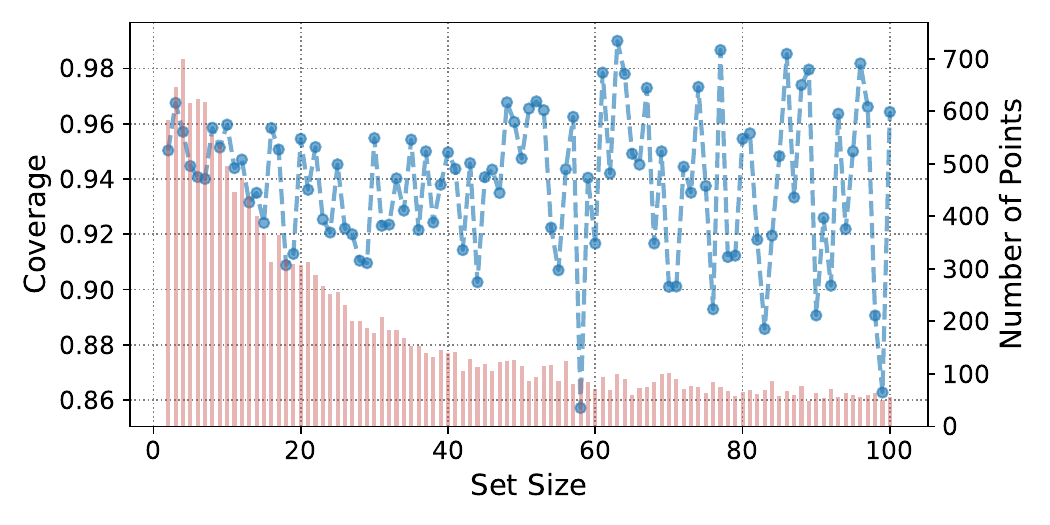}
        \caption{Conditional coverage for OPT$_\text{(350M)}$ on Language Modelling.}
        \label{subfig:tratified-coverage-opt}
    \end{subfigure}
    \begin{subfigure}[t]{\columnwidth}
        \centering
        \includegraphics[width=0.9975\columnwidth]{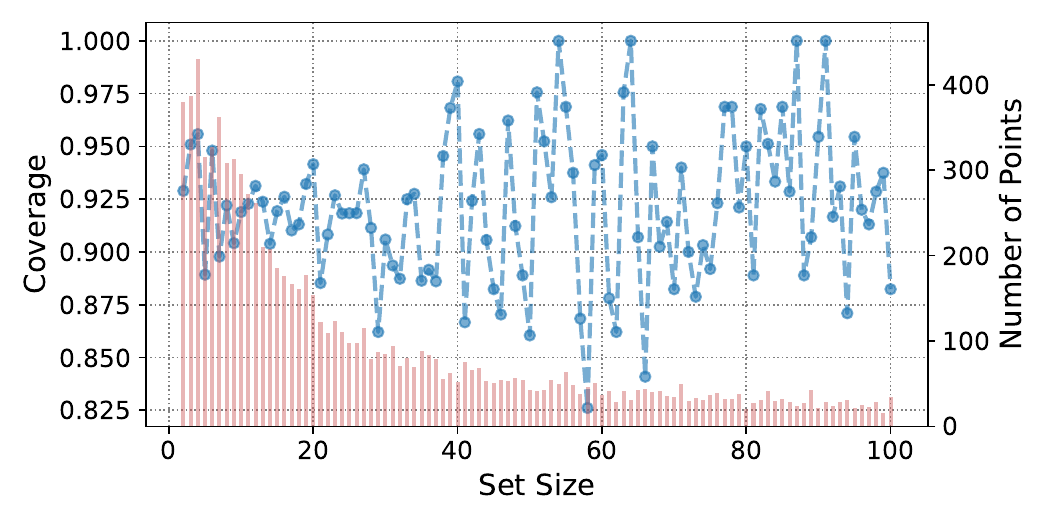}
        \caption{Conditional coverage for OPT$_\text{(1.3B)}$ on Language Modelling.}
        \label{subfig:stratified-coverage-opt-large}
    \end{subfigure}
    \caption{Additional conditional coverage plots for the MT and LM dataset using our non-exchangeable conformal prediction method, aggregating predictions by prediction set size. The blue curve shows the conditional coverage per bin, whereas red bars show the number of predictions per bin. For \cref{subfig:tratified-coverage-opt,subfig:stratified-coverage-opt-large}, we zoom in on the prediction set sizes from $1$ and $100$.}\label{fig:additional-conditional-coverage}
\end{figure}

We show additional plots illustrating the coverage per set size-bins in \cref{fig:additional-conditional-coverage}. 
We can see the counterparts for \cref{fig:conditional-coverage} using the larger M2M100$_\text{(1.2B)}$ model in \cref{subfig:stratified-coverage-deen-m2m00large,subfig:stratified-coverage-jaen-m2m00large}:
Instead of leveling off like for the smaller model, most prediction set sizes are either in a very small range or in a size of a few ten thousand.
In \cref{subfig:tratified-coverage-opt,subfig:stratified-coverage-opt-large}, we show similar plots for the two different OPT model sizes. 
Since in both cases, most prediction set sizes are rather small, we zoom in on the the sizes from $1$ to $100$. Here, we can observe a similar behavior to the smaller M2M100$_\text{(400m)}$, gradually leveling off. 
We do not show similar plots for other distance metrics as they show similar trends.

\subsection{Impact of Coverage Threshold and Neighborhood Size Choice}\label{app:ablations}

In this section, we present experiments surrounding the two most pivotal parameters of our method: The desired confidence level $\alpha$, as well as the number of neighbors.

\begin{table}[htb!]
    \centering
    \resizebox{0.475\textwidth}{!}{
    \renewcommand{\arraystretch}{1.4}
        \begin{tabular}{@{}llrrrr@{}}
        \toprule[0.05cm]
         & $\alpha$ & \% \textsc{Cov.} & $\varnothing$ \textsc{Width} $\downarrow$ &  \textsc{Scc} $\uparrow$ & \textsc{Ecg} $\downarrow$ \\
         \midrule
        \multirow{9}{*}{\rotatebox[origin=c]{90}{M2M100$_\text{(400M)}$ / de $ \rightarrow $ en}} & $0.1$ & $0.9442$ & $0.31$ & $0.8702$ & $0.0011$ \\
        & $0.2$ & $0.8767$ & $0.18$ & $0.7906$ & $8.63 \times 10^{-5}$ \\
        & $0.3$ & $0.7963$ & $0.12$ & $0$ & $0.0016$ \\
        & $0.4$ & $0.7058$ & $0.09$ & $0.1393$ & $0.0082$ \\
        & $0.5$ & $0.6081$ & $0.07$ & $ 0.2836$ & $0.0055$ \\
        & $0.6$ & $0.5017$ & $0.06$ & $0.1393$ & $0.0082$ \\
        & $0.7$ & $0.3896$ & $0.05$ & $0$ & $0.0091$ \\
        & $0.8$ & $0.2800$ & $0.05$ & $0$ & $0.0090$ \\
        & $0.9$ & $0.1762$ & $0.04$ & $0$ & $0.0071$ \\
        \midrule
        \multirow{9}{*}{\rotatebox[origin=c]{90}{M2M100$_\text{(400M)}$ / ja $ \rightarrow $ en}} & $0.1$ & $0.7453$ & $0.15$ & $0.3080$ & $0.1511$ \\
        & $0.2$ & $0.5579$ & $0.07$ & $ 0.2728$ & $0.2446$ \\
        & $0.3$ & $0.4277$ & $0.04$ & $0.2770$ & $0.2779$ \\
        & $0.4$ & $0.3438$ & $0.03$ & $0.1212$ & $0.2438$\\
        & $0.5$ & $0.2749$ & $0.03$ & $0.0455$ & $0.1883$ \\
        & $0.6$ & $0.2175$ & $0.02$ & $0$ & $0.1207$\\
        & $0.7$ & $0.1685$ & $0.02$ & $0$ & $0.0560$ \\
        & $0.8$ & $0.1309$ & $0.01$ & $0$ & $0.0117$ \\
        & $0.9$ & $0.0989$ & $0.02$ & $0$ & $0.0099$\\
        \midrule
        \multirow{9}{*}{\rotatebox[origin=c]{90}{OPT$_\text{(350M)}$ / \textsc{OpenWebText}}} & $0.1$ & $0.9460$ & $0.26$ & $0.8$ & $1.85 \times 10^{-5}$ \\
        & $0.2$ & $0.8937$ & $0.16$ & $0.8$ & $0$ \\
        & $0.3$ & $0.8392$ & $0.10$ & $0.5$ & $8.74 \times 10^{-6}$ \\
        & $0.4$ & $0.7782$ & $0.08$ & $0.6667$ & $0$ \\
        & $0.5$ & $0.7171$ & $0.06$ & $0$ & $1.19 \times 10^{-5}$ \\
        & $0.6$ & $0.6559$ & $0.06$ & $0.6033$ & $0$ \\
        & $0.7$ & $0.5945$ & $0.05$ & $0$ & $8.21 \times 10^{-6}$ \\
        & $0.8$ & $0.5349$ & $0.05$ & $0.4462$ & $0$ \\
        & $0.9$ & $0.4757$ & $0.05$ & $0.3580$ & $0$ \\
        \bottomrule[0.05cm]
    \end{tabular}%
    }
    \caption{Results for different values of $\alpha$ using different models and datasets.}
    \label{tab:alpha-ablation}
\end{table}

\begin{table}[htb!]
    \centering
    \resizebox{0.475\textwidth}{!}{
    \renewcommand{\arraystretch}{1.4}
        \begin{tabular}{@{}llrrrr@{}}
        \toprule[0.05cm]
         & $K$ & \% \textsc{Cov.} & $\varnothing$ \textsc{Width} $\downarrow$ &  \textsc{Scc} $\uparrow$ & \textsc{Ecg} $\downarrow$ \\
         \midrule
        \multirow{9}{*}{\rotatebox[origin=c]{90}{\hspace{0.75cm} M2M100$_\text{(400M)}$ / de $ \rightarrow $ en}} & $10$ & $0.9923$ & $0.39$ & $0.9728$ & $0$ \\
        & $25$ & $0.9563$ & $0.37$ & $0.8877$ & $0.0011$ \\
        & $50$ & $0.9504$ & $0.32$ & $0.8870$ &  $0.0006$\\
        & $75$ & $0.9444$ & $0.32$ & $0.8641$ & $0.0014$ \\
        & $100$ & $0.9442$ & $0.31$ & $0.8702$ & $0.0011$ \\
        & $200$ & $0.9422$ & $0.31$ & $0.8125$ & $0.0016$ \\
        & $300$ & $0.9404$ & $0.31$ & $0.8483$ & $0.0019$ \\
        & $500$ & $0.9389$ & $0.31$ & $0.8214$ & $0.0023$ \\
        \midrule
        \multirow{9}{*}{\rotatebox[origin=c]{90}{\hspace{0.75cm} M2M100$_\text{(400M)}$ / ja $ \rightarrow $ en}} & $10$ & $0.8013$ & $0.17$ & $0.2995$ & $0.1606$ \\
        & $25$ & $0.7353$ & $0.17$ & $0.2994$ & $0.1438$ \\
        & $50$ & $0.7540$ & $0.17$ & $0.3023$ & $0.1603$ \\
        & $75$ & $0.7368$ & $0.16$ & $0.3019$ & $0.1603$ \\
        & $100$ & $0.7453$ & $0.15$ & $0.3072$ & $0.1529$ \\
        & $200$ & $0.7295$ & $0.14$ & $0.2938$ & $0.1787$ \\
        & $300$ & $0.7192$ & $0.13$ & $0.2948$ & $0.1788$ \\
        & $500$ & $0.7110$ & $0.13$ & $0.2756$ & $0.1867$ \\
        \midrule
        \multirow{9}{*}{\rotatebox[origin=c]{90}{\hspace{0.75cm} OPT$_\text{(350M)}$ / \textsc{OpenWebText}}} & $10$ & $0.9438$ & $0.35$ & $0.8824$ & $0.0019$ \\
        & $25$ & $0.9522$ & $0.33$ & $0.8333$ & $2.06 \times 10^{-5}$ \\
        & $50$ & $0.9442$ & $0.27$ & $0$ & $1.86 \times 10^{-5}$\\
        & $75$ & $0.9477$ & $0.27$ & $0.8$ & $1.03 \times 10^{-5}$ \\
        & $100$ & $0.9460$ & $0.26$ & $0.8$ & $1.86 \times 10^{-5}$ \\
        & $200$ & $0.9487$ & $0.28$ & $0.8571$ & $6.20 \times 10^{-5}$ \\
        & $300$ & $0.9500$ & $0.28$ & $0.8181$ & $1.86 \times 10^{-5}$ \\
        & $500$ & $0.9508$ & $0.29$ & $0.8181$ & $1.86 \times 10^{-5}$ \\
        \bottomrule[0.05cm]
    \end{tabular}%
    }
    \caption{Results for different neighborhood sizes $K$ using different models and datasets.}
    \label{tab:neighbor-ablation}
\end{table}

\paragraph{Coverage Threshold.} In \cref{tab:alpha-ablation}, we investigate the impact of different values on $\alpha$ on our evaluation metrics. 
We show that the increase in $\alpha$ does indeed produce the expected decrease in coverage, however with a certain degree of overcoverage for the de $ \rightarrow $ en MT and the LM task.
The loss in coverage always goes hand in hand with a decrease in the average prediction set width as well, as the model can allow itself to produce tighter prediction sets at the cost of higher miscoverage. 
As this also produces bin in which all contained instances are uncovered, this produces zero values for the SCC, while we cannot discern clear trends for the ECG.

\paragraph{Neighborhood Size.} In \cref{tab:neighbor-ablation}, we vary the effect of the chosen neighborhood size (with $100$ being the value we use in our main experiments). 
We make the following, interesting observations: Coverage on the MT task seems to decrease with an increase in the neighborhood size as prediction set widths get smaller on average, with a neighborhood size around $100$ striking a balance between coverage, width, computational cost and SCC / ECG.
For LM, coverage seems to be mostly constant, with prediction set width hitting an inflection point for $100$ neighbors.
We speculate that initially there might be a benefit to considering more neighbors to calibrate $\hat{q}$, but that considering too large neighborhoods might introduce extra noise. 
While we found $100$ to be a solid choice for the purpose of our experiments, we leave more principled ways to determine the neighborhood size to future work.

\subsection{Algorithm}\label{app:algorithm}

\begin{algorithm}
\caption{Non-exchangeable Conformal Language Generation with Nearest Neighbors}\label{alg:cap}
\begin{algorithmic}
\Require Sequence $\bx^{(i)}$, model $f_{\btheta}$, datastore $\text{DS}(\cdot)$ with model activations collected from held-out set, temperature $\tau$
\State
\While{generating}
    \LineComment{1. Extract latent encoding for current input}
    \State $\bz^{(i)}_t \gets f_{\btheta}(\bx_t)$\\
    \LineComment{2. Retrieve $K$ neighbors \& non-conformity scores}
    \State $\{(\bz_1, s_1), \ldots (\bz_K,  s_K)\} \gets \text{DS}
    (\bz_t)$\\
    \LineComment{3. Compute weights $w_k$ and normalize}
    \State $w_k \gets\ \exp(-||\bz_t^* - {\bz_k}||^2_2\ /\ \tau)$
    \State $\tilde{w}_k\ \leftarrow\ w_k / (1 + \sum_{k=1}^K w_k)$\\
    
    \LineComment{4. Find quantile $\hat{q}$}
    \State $\hat{q} \gets\ \inf \{q\ | \sum_{i=1}^N \tilde{w}_i \mathbf{1}\{s_i\le q \} \ge 1 - \alpha \}$\\

    \LineComment{5. Create prediction set}
    \State $\hat{c} \gets\ \sup \{c^\prime | \sum_{j=1}^{c^\prime} p_{\btheta}(y=\pi(j)|\bx^*) < \hat{q}\} + 1$
    \State $\mathcal{C}(\bx^*) \gets\ \{\pi(1), \ldots, \pi(\hat{c})\}$\\

    \LineComment{6. Generate next token}
    \State $y_t\ \leftarrow\ \text{generate}(\mathcal{C}(\bx^*))$
    
\EndWhile

\end{algorithmic}
\end{algorithm}

We show the algorithm that was schematically depicted in \cref{fig:schematic} in pseudo-code in \cref{alg:cap}.
It mostly requires that we have pre-generated a datastore of latent representations of the model on a held-out set along with their non-conformity scores (in our case, using the score defined in \ref{eq:adaptive-non-conformity-score} and the FAISS \citep{johnson2019billion} as the datastore architecture).
Furthermore, we need to have determined an appropriate value for the temperature $\tau$ in advance (see \cref{app:temperature-search}).
Then, the algorithm involves the following steps:

\begin{enumerate}
    \item Extract the latent encoding for the current time step $\bz_t$ from the model. Even though different options are imaginable, we utilize the activations of the uppermost layer.
    \item Retrieve $K$ neighbors and their corresponding non-conformity scores from the datastore.
    \item Compute the weights $w_k$ based on the squared $l_2$ distance between $\bz_t$ and its neighbors in the datastore and normalize the weights to obtain $\tilde{w}_k$.
    \item Use \cref{eq:non-exchangeable-quantile} to find the quantile $\hat{q}$.
    \item Use $\hat{q}$ to create prediction sets, for instance the adaptive prediction sets defined in \cref{eq:adaptive-prediction-set}.
    \item Finally, generate the new token $y_t$ by sampling from the prediction set. 
\end{enumerate}

The main computational bottleneck of this algorithm is the retrieval process that fetches the closest neighbors from the datastore during every generation step.
However, while not explored further in this work, there are some potential avenues to reduce this load:
On the one hand, works such as \citet{he2021efficient, henrique2022chunk} have demonstrated ways to reduce the computational load of $k$-NN based approaches.
On other hand, we treat the number of neighbors  $K$ fixed during every generation step.
However, it seems intuitive that the number of neighbors necessary to create good prediction sets would not be the same for all tokens. 
Future research could explore setting $K$ dynamically during every time step, thus reducing the overall slowdown.

\subsection{Environmental Impact}\label{app:environmental-impact}

We track the carbon emissions produced by this work using the \texttt{codecarbon} tracking tool \citep{codecarbon, lacoste2019quantifying, lottick2019nergy}. 
The carbon efficiency was estimated to be $0.12$ kgCO$_2$eq / kWh. 
159.5 hours of computation were performed on a NVIDIA RTX A6000.
Total emissions are estimated to be $6.99$ kgCo2eq.
All of these values are upper bound including debugging as well as failed or redundant runs, and thus any replication of results will likely be shorter and incur fewer carbon emissions.

\subsection{Qualitative Analysis}\label{app:qualitative-analysis}

In \cref{tab:qualitative-samples1,tab:qualitative-samples2,tab:qualitative-samples3,tab:qualitative-samples4}, we show a few samples from the different methods on our used datasets.
We could observe some general patterns from the given and further samples:
For the translation task, our method and other sampling techniques seem to sometimes prefer literal but unfitting translations.
As expected for text generation, greedy sampling produces the worst translations, often getting stuck in loops, while beam search produces incoherent outputs.
Interestingly, non-exchangeable conformal sampling and conformal sampling seem to often produce similar sampling before diverging.

\begin{table*}
    \centering
    \resizebox{0.975\textwidth}{!}{
    \renewcommand{\arraystretch}{1.4}
        \begin{tabular}{@{}p{0.1\linewidth}p{0.92\linewidth}@{}}
            \toprule
            Input & Außerdem funktioniert der Brandschutz am Übergang zum Bahnhof unter dem Terminal noch nicht zufriedenstellend. Die Probleme hatten das Projekt schon vor drei Jahren in eine weitere Krise geführt. Dieses Mal geht es um Rauchschürzen, die bei einem Brand wie eine Leinwand herunterfahren, damit der Rauch sich nicht ausbreitet. Sie fahren zu weit herunter. 36 Motoren müssen deshalb ausgetauscht werden. Dann soll es Anfang des Jahres neue Versuche mit Heißgasrauch geben. \\
            Reference & Also, the fire protection at the passage to the station under the Terminal is not satisfactory yet. There had already been problems leading the project into further crisis three years ago. This time it's about smoke barriers, which descend like a canvas during a fire, so that smoke does not spread. They go too far down. Therefore, 36 engines have to be exchanged. There will be new testing with hot gas smoke at the beginning of the year. \\
            \midrule
            Beam Search & In addition, the fire protection at the transition to the station under the terminal is not yet satisfying. The problems had led the project to another crisis three years ago. This time it is about smoke cuts that fall down in a fire like a wall so that the smoke does not spread. They run too far down. 36 engines therefore need to be exchanged. Then there will be new attempts with hot gas smoke at the beginning of the year.\\
            Greedy & In addition, the fire protection at the transition to the station under the terminal is still not satisfying. The problems had led the project to another crisis three years ago. This time it is about smoke shakes that fall down in a fire like a wall so that the smoke does not spread. They run too far down. 36 engines therefore need to be exchanged. Then it is expected to be new attempts with hot gas smoke at the beginning of the year. \\
            Nucleus Sampling & In addition, the fire protection at the transition to the station under the terminal is not yet satisfying. The problems had led the project to another crisis three years ago. This time it is about smoke shakes that fall down in a fire like a wall so that the smoke does not spread. They run too far down. 36 engines therefore need to be exchanged. Then it is expected to be new attempts with hot gas smoke at the beginning of the year.\\
            Top-$k$ Sampling & In addition, the fire protection at the transition to the station under the terminal is not yet satisfying. The problems had led the project to another crisis three years ago. This time it is about smoke shakes that fall down in a fire like a wall so that the smoke does not spread. They run too far down. 36 engines therefore need to be exchanged. Then it is expected to be new attempts with hot gas smoke at the beginning of the year.\\
            Conf. Sampling & In addition, the fire protection at the transition to the station under the terminal is not yet satisfying. The problems had led the project to another crisis three years ago. This time it is about smoke shakes that fall down in a fire like a wall so that the smoke does not spread. They run too far down. 36 engines therefore need to be exchanged. Then it is expected to be new attempts with hot gas smoke at the beginning of the year. \\
            Non-Ex. CS & In addition, fire protection at the transition to the station under the terminal is still not satisfying. The problems had led the project to another crisis three years ago. This time it is about smoke cuts that fall down in a fire like a wall so that the smoke does not spread. They run too far down. 36 engines therefore need to be exchanged. Then there will be new attempts with hot gas smoke at the beginning of the year. \\
            \bottomrule
        \end{tabular}%
    }
    \caption{Samples from M2M100$_\text{(400M)}$ on the de $\rightarrow$ en translation task.}\label{tab:qualitative-samples1}
\end{table*}

\begin{table*}
    \centering
    \resizebox{0.975\textwidth}{!}{
    \renewcommand{\arraystretch}{1.4}
        \begin{tabular}{@{}p{0.1\linewidth}p{0.92\linewidth}@{}}
            \toprule
            Input & Angesichts der aufgeladenen Stimmung riefen am Freitag sogar die Bischöfe der anglikanischen Kirche zur Zurückhaltung auf. "Wir sollten miteinander mit Respekt sprechen"", hieß es in einer Erklärung. "Und wir sollten auch zuhören". \\
            Reference & In view of the charged mood, even bishops of the Anglican Church called for restraint on Friday. "We should speak to others with respect. And we should also listen." \\
            \midrule
            Beam Search & In view of the loaded mood, even the Bishops of the Anglican Church called for refusal on Friday. ”We should speak with respect to each other,” it said in a statement. ”And we should also listen.” \\
            Greedy & In light of the loaded mood, even the Bishops of the Anglican Church called for refusal on Friday. ”We should speak with respect to each other,” the statement said. ”And we should listen.”\\
            Nucleus Sampling & In light of the loaded mood, even the Bishops of the Anglican Church called for refusal on Friday. ”We should speak with respect to each other,” the statement said. ”And we should listen.” \\
            Top-$k$ Sampling & In light of the loaded mood, even the Bishops of the Anglican Church called for refusal on Friday. ”We should speak with respect to each other,” the statement said. ”And we should listen.” \\
            Conf. Sampling & In light of the loaded mood, even the Bishops of the Anglican Church called for refusal on Friday. ”We should speak with respect to each other,” the statement said. ”And we should listen.” \\
            Non-Ex. CS & In light of the burden of mood, even the Bishops of the Anglican Church called for a rejection on Friday. ”We should speak with respect to each other,” the statement said. ”And we should listen.”\\
            \bottomrule
        \end{tabular}%
    }
    \caption{Samples from M2M100$_\text{(400M)}$ on the de $\rightarrow$ en translation task.}\label{tab:qualitative-samples2}
\end{table*}

\begin{table*}
    \centering
    \small
    \resizebox{0.995\textwidth}{!}{
    \renewcommand{\arraystretch}{1.4}
        \begin{tabular}{@{}p{0.1\linewidth}p{0.92\linewidth} @{}}
            \toprule
            Prompt & TORONTO — Social media posts encouraging people to visit southern Ontario's wineries by bicycle has prompted a regional public transit service to apologize for the "misunderstanding."

In May 10 Facebook and Twitter postings, GO Transit said \\
            Reference &  "Niagara's wineries are best enjoyed responsibly and on two wheels."

The posting encouraged people to take their bicycle on the GO for "a cycling weekend adventure in Ontario's wine country" in a promotion for its seasonal service to Niagara Region.

That prompted numerous replies on social media that suggested Go Transit was endorsing mixing cycling and drinking. \\
            \midrule
            Beam Search & , encouraged people to visit wineries in southern Ontario by bicycle. "We apologize for the misunderstanding," GO Transit said in a statement. "We are aware of the posts and are working with our partners to address the issue." GO Transit said the posts were not intended to encourage people to visit wineries in southern Ontario. "We are aware of the posts and are working with our partners to address the issue," GO Transit said in a statement. \\
            Greedy & , encouraged people to visit wineries in the region. "We are sorry for the misunderstanding," GO Transit said in a statement. "We are working with the local community to address this issue and apologize for any inconvenience caused." The posts were posted by a person who was not identified. The posts were deleted. The posts were posted by a person who was not identified. (Submitted by GO Transit) The posts were posted by a person who was not identified. The posts were posted by a person who was not identified. The posts were posted by a person who was not identified. The posts were posted by a person who was not identified. The posts were posted by a person who was not identified. The posts were posted by a person who was not identified. The posts were posted by a person who was not identified. The posts were posted by a person who was not identified. The posts were posted by a person who was \\
            Nucleus Sampling & it might suspend riders from Ontario's wineries after several people raised concerns about the company's planned stop at Harney and Sons Winery. The message went out in English and Spanish and was circulated with other posts throughout the province. But GO Transit told the Globe and Mail Wednesday it was disappointed that the posts did not mention the wineries. Trains currently leave each of their stops from Toronto's east end to south of Toronto's west end, and at certain times throughout the day from each of those stops to a number of wineries. Go Transit said it has apologized and is changing its approach to its route network. The transit agency is also changing its system to include fewer stopping locations, including stops at wineries. Ontario's regional public transit services are to be revamped to include a more regional approach. CTV News Canada CBC News Radio has reached out to GO Transit for comment. \\
            Top-$k$ Sampling & in a post that "there is a need to encourage people to visit wineries in southern Ontario." "We apologize to the public at large for this misunderstanding," the post said. "We are committed to educating the public as we work to make GO Transit safer and efficient," the statement said. The posts, which appeared on the GO Transit website, also encouraged people to drive through the region. It's unclear if the posts were meant to be taken seriously or if they were taken as a joke about driving through southern Ontario. "In the past we have encouraged the public to visit wineries in the region and have since been very pleased with this response," the transit service wrote on the website. "While we are not able to guarantee that there will be no traffic problems, we would like to encourage people to visit wineries in the region and to take advantage of these unique locations." GO Transit said it's committed to improving its operations" \\
            Conf. Sampling & in a post that "there is a reason it's called a cycling route." But in a response to a series of responses to a complaint to the TransLink Board of Directors from the provincial government, the Vancouver-based service says it is wrong to call the route a cycling route. In fact, GO says the transit service was referring to a cycling route. "GO says it was looking into the possible error — one that many people are probably aware of," said the statement. The B.C. government is also investigating the posts. Earlier this week a Vancouver woman was struck and killed after she was run over in a drive-by shooting on her way to a restaurant by the Vancouver Coastal Health NHS. RCMP are still investigating the killing. \\
            Non-Ex. CS & in a post that "there is a reason it's called a cycling route." But in a response to a series of responses to a complaint to the TransLink Board of Directors from the provincial government, the Vancouver-based service says it is wrong to call the route a cycling route. In fact, GO says the transit service was referring to a cycling route. ""GO says it was looking into the possible error — one that many people are probably aware of," said the statement. "We know we were mistaken. "In addition, in the years since we initially started the bike program, we have had many comments that this bike route is not actually a cycling route." The statement continues to explain the changes it made after the original post, including making it clearer that while the route is a cycleway, it is not considered a commuter route \\
            \bottomrule
        \end{tabular}%
    }
    \caption{Samples from OPT$_\text{(350M)}$on OpenWebText.}\label{tab:qualitative-samples3}
\end{table*}
\begin{table*}
    \centering
    \small
    \resizebox{0.995\textwidth}{!}{
    \renewcommand{\arraystretch}{1.4}
        \begin{tabular}{p{0.085\linewidth}p{0.92\linewidth} }
            \toprule
            Prompt & PARIS -- Pepe is set to become Paris Saint-Germain's first summer signing once his Real Madrid contract comes to an end, a source close to the French capital outfit told ESPN FC.

            The Portugal international is \\
            Reference & still on Confederations Cup duty and will be until the third-placed playoff on Sunday, just two days before PSG coach Unai Emery and his non-international players report to Camp des Loges for preseason training, but the 34-year-old is close to signing a contract.

According to the source, Pepe has been in talks with Les Parisiens for some time over a potential move to Parc des Princes -- before and after compatriot Antero Henrique's arrival as sporting director.

However, it is mainly because of the former Porto man that this deal has been pushed to the brink of completion after the Portuguese transfer guru and Emery agreed the squad require greater experience and a stronger winning mentality. \\
            \midrule
            Beam Search & set to sign a two-year deal with the French club, which is expected to be completed by the end of the month, the source said. Pepe, who joined Real Madrid from Real Sociedad in the summer of 2014, has been linked with a move to Paris Saint-Germain since the summer of 2015. The 22-year-old has been a key player for Real Madrid since joining from Sociedad in the summer of 2014. Pepe has scored five goals and provided three assists for Real Madrid this season. \\
            Greedy & set to sign a two-year deal with the French club, which is expected to be completed in the coming days, the source said. The source said that the club is in talks with the player's agent, who is expected to be in Paris on Monday to finalize the deal. The source said that the player is expected to sign a contract with the club until the end of the season. The source said that the player is expected to sign a contract with the club until the end of the season. The source said that the player is expected to sign a contract with the club until the end of the season. The source said that the player is expected to sign a contract with the club until the end of the season. The source said that the player is expected to sign a contract with the club until the end of the season. The source said that the player is expected to sign \\
            Nucleus Sampling & in the final 12 months of his European Super League contract and has already signed a new contract with the club. Spurs and Liverpool - along with Real -- signed the 20-year-old last summer -- their first signings from the group that includes Everton, Arsenal, West Ham, Manchester United and Chelsea. He is the fifth player to join the club this summer, with Nuno Espirito Santo joining on the half-yearly expiry, with Real joining along with Liverpool, Chelsea and Tottenham on the year-long contract. With Pepe, PSG were looking to upgrade at the back with Kalidou Koulibaly a target for PSG, while PSG also made a small but interesting splash in the form of the arrival of Marco Verratti from RB Leipzig. Pepe has started all of the club's matches so far this season and has scored and assisted as many goals as has been possible due to the lack of physicality of his opponents in the midfield. He has become the club's third\\
            Top-$k$ Sampling & a free agent on July 1, meaning PSG will likely look to fill some of the gaping openings in its defensive front three. ""e is the most likely to join, but there's no official news yet," the source said. "But the deal will happen once it's signed." Paris Saint-Germain's Pepe (centre) talks to Paris Saint-Germain's head scout, Bruno Salles, during training. Photo: AFP The 23-year-old was part of PSG's academy system at the start of the season, and has already started to make strides in his development after a series of impressive displays. The club has already made two signings this summer - signing former Manchester United goalkeeper Edinson Cavani on loan and loanee midfielder Nabil Bentaleb from Ajax. But it has also seen its recruitment strategy change in the wake of Cristiano Ronaldo's exit to Juventus, with Neymar returning to Paris as its new manager. PSG has also made some big signings this summer, with the likes of Neymar, Kylian Mbappe, Raphael Varane,\\
            Conf. Sampling & a free agent on July 1 -- meaning PSG cannot sign him until the club's financial regulations are in place. The 21-year-old began his career at PSG's youth setup between 2006 and 2007 under former coach Julien Laurens. He went on to start four appearances in France's U-21 World Cup qualifying campaign in 2012-13, but LaLiga action dried up due to a lack of reinforcements. "There are no rumours," a Paris Saint-Germain source told ESPN FC. "Pepe is a PSG player and he has signed for us. There is nothing that we can say about his future." PSG will pay close attention to the recruitment of new signing Raphael Varane, whose contract is due to expire in the summer of 2016." \\
            Non-Ex. CS & a free agent on July 1, meaning PSG cannot sign him until the club's summer transfer window opens on August 3. The story said that PSG has tried to win a signature of the 26-year-old before he will close his door on Real, but it appears the club is now ready to make another deal. Portugal international Pepe's departure from Real is very much expected and PSG must now move for another one of its first-team players, after two disappointing season. The club failed to earn a top four finish in 2017/18. While PSG's first-team squad included several transfers ahead of the 2020/21 season, Pepe's departure would give the French club just enough options to deal with Real's spending. There was also the possibility of a deal for Brazilian international winger Angel di Maria. But he never signed with PSG after the club's financial difficulties with UEFA's financial fair play framework. Real's budget in 2018/19 was over €7M to fund Cristiano Ronaldo's exit from \\
            \bottomrule
        \end{tabular}%
    }
    \caption{Samples from OPT$_\text{(350M)}$ on OpenWebText.}\label{tab:qualitative-samples4}
\end{table*}

\end{document}